\documentclass[%
reprint,
nofootinbib,
 amsmath,amssymb,
aps,
superscriptaddress,
showkeys,
longbibliography
]{revtex4-1}

\usepackage{enumitem}

\usepackage{xr}
\usepackage{tabularx}
\usepackage{graphicx}
\usepackage{dcolumn}
\usepackage{bm}
\usepackage{microtype}
\usepackage{gensymb}
\usepackage{url}
\usepackage[breaklinks, hidelinks, colorlinks=true,linkcolor=blue,citecolor=black]{hyperref}
\usepackage{color, colortbl}
\usepackage[table,xcdraw]{xcolor}
\usepackage{booktabs}
\usepackage{makecell}
\usepackage{multirow}
\usepackage{placeins}

\usepackage[table]{xcolor}

\makeatletter
\renewcommand\frontmatter@abstractwidth{\dimexpr0.9\textwidth\relax}
\makeatother

\usepackage{algorithm}
\usepackage{algpseudocode}

\makeatletter
\newcommand*{\addFileDependency}[1]{
  \typeout{(#1)}
  \@addtofilelist{#1}
  \IfFileExists{#1}{}{\typeout{No file #1.}}
}
\makeatother

\makeatletter
\renewcommand\subparagraph{\@startsection{subparagraph}{5}{\parindent}%
    {3.25ex \@plus1ex \@minus .2ex}%
    {-1em}%
    {\normalfont\normalsize\bfseries}}
\makeatother

\begin{document}

\preprint{1}

\title{Evaluation of Blood Vessel Segmentation Methods on Hard-to-Detect Vascular Structures}%

\author{João P. Parella}
\affiliation{Department of Computer Science, Federal University of S\~ao Carlos, S\~ao Carlos, SP, Brazil}

\author{Matheus V. da Silva}
\affiliation{Department of Computer Science, Federal University of S\~ao Carlos, S\~ao Carlos, SP, Brazil}

\author{Cesar H. Comin}
\email[Corresponding author: ]{comin@ufscar.br}
\affiliation{Department of Computer Science, Federal University of S\~ao Carlos, S\~ao Carlos, SP, Brazil}

\date{\today}

\begin{abstract}

\noindent{\bf Background and Objective:} 
Due to the intricate structure of vascular trees, minor segmentation errors can significantly alter connectivity patterns and increase variability in extracted morphological properties. Global metrics such as the Dice coefficient, precision, and recall often overlook inaccuracies in specific regions of a sample. To address this, we define a Local Vessel Salience (LVS) index to quantify the difficulty of identifying specific vessel segments. This index is used to evaluate the performance of 16 segmentation methods across six widely used 2D datasets.
{\bf Methods:} The LVS index is calculated for each vessel pixel by comparing local vessel intensity against the surrounding background. We introduce a metric termed mean Low-Salience Recall (mLSR) to quantify how effectively algorithms recover hard-to-detect vessels. Furthermore, we propose a proof-of-concept data augmentation procedure guided by the LVS index aimed at improving neural network segmentation performance.
{\bf Results:} Our findings demonstrate that segmentation performance strongly correlates with LVS, revealing systematic errors in vessels with low salience. The mLSR across all evaluated methods was significantly lower than standard recall values, with a typical performance decrease of 20 percentage points. In benchmark datasets such as DRIVE and OCTA-500, mLSR values for the top-performing methods were approximately 58\%. 
{\bf Conclusions:} The developed methodology provides a quantitative basis for the design of segmentation algorithms with improved sensitivity to hard-to-detect vessels and enhanced capabilities for preserving vascular connectivity.

\end{abstract}


\maketitle
\thispagestyle{plain}

\section{Introduction}
\label{sec:introduction}

Identifying blood vessels in digital images is critical for clinical diagnosis \cite{mookiah2021review,eladawi2018early,sangeethaa2018intelligent,li2021blood,almotiri2018multi,nair2020blood} and for obtaining precise measurements to support vascular research \cite{roda2021blood,ouellette2020vascular,wong2019blood,dolati2015pre}. Blood vessels exhibit diverse morphologies and form complex, interconnected structures that typically span an entire tissue sample. Segmenting these vascular networks is essential for downstream analyses, including branching point characterization, tortuosity quantification, and blood flow simulation \cite{li2022human,ouellette2020vascular,fraz2012blood}. A standard processing pipeline involves vasculature segmentation, followed by the extraction of medial axes to create a graph-based representation of the network \cite{paetzold2021whole,freitas2022unbiased}.

Numerous methodologies have been developed for blood vessel identification in digital images \cite{mookiah2021review,cervantes2023comprehensive}. Recent techniques utilize Convolutional Neural Networks (CNNs) trained on manually annotated samples to segment vessels within a dataset \cite{chen2021retinal}. These approaches commonly evaluate performance using accuracy and the Dice coefficient, which quantify overall segmentation quality. Boundary distance metrics \cite{schaap2009standardized,van2008averaging} and the recently developed clDice metric \cite{shit2021cldice}, which assesses connectivity preservation, are also employed.

\begin{figure}
    \centering
    \includegraphics[width=\linewidth]{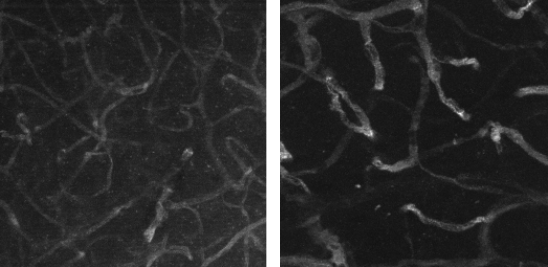}
    \caption{Two fluorescence microscopy images of the mouse cortex~\cite{freitas2022unbiased} showing blood vessels with varying salience.}
    \label{f:examples}
\end{figure}

However, significant variations in saliency between vessels and the background are common, as illustrated in Figure~\ref{f:examples}. While global metrics and topology-aware measures like clDice evaluate general quality, they fail to isolate an algorithm's performance on hard-to-detect, low-salience structures. Because standard metrics aggregate performance across the entire image, predominant high-contrast vessels heavily bias the final score, masking localized failures. Small discontinuities along low-salience vessels can fundamentally alter the connectivity of the extracted network, yet their impact is often diluted in global metrics.

To address this limitation, we introduce the Local Vessel Salience (LVS) index for each vascular pixel. The index quantifies the expected difficulty of segmenting a specific vessel region by comparing local vessel intensity with the immediate background. We subsequently use this index to define the mean Low-Salience Recall (mLSR), a metric that quantifies the average recall within challenging vessel segments\footnote{Code available at \url{https://github.com/jpparella/vessel_salience}}. The mLSR provides critical insights into segmentation performance for tasks sensitive to vascular connectivity. Our results demonstrate that for all analyzed methods, the mLSR is significantly lower than standard recall, indicating a consistent failure to recover low-salience vessels.

Furthermore, we present a salience-based data augmentation technique designed to improve network robustness in these regions. The augmentation discourages the network's reliance on local contrast by artificially degrading the LVS of specific vessel segments and simulating realistic discontinuities during training. This encourages the model to infer vessel presence based on the global continuity of the vascular tree. We demonstrate that this targeted approach is promising in scenarios with limited training data.

The remainder of this paper is organized as follows: Section~\ref{sec:related} discusses related work; Section~\ref{s:lvs} defines the LVS index; Section~\ref{s:lsrecall} and Section~\ref{s:lsaug} present the mLSR score and the augmentation procedure, respectively; Section~\ref{s:results} details the experimental results; and Section~\ref{s:conclusion} concludes the study.

\section{Related Works}
\label{sec:related}

The segmentation of interconnected curvilinear structures is a persistent challenge in computer vision. Significant research has been dedicated to extracting road networks from aerial imagery \cite{mosinska2018beyond,citraro2020towards,vasu2020topoal}, identifying neuronal dendrites and axons \cite{hu2019topology,funke2018large,scheffer2020connectome}, and reconstructing complex vascular networks \cite{gupta2024topology,hu2021topology}. The objective of this study is not to introduce a novel segmentation or graph-generation algorithm, but rather to propose a metric that evaluates existing methods through the lens of local difficulty.

Various metrics have been developed to assess blood vessel identification performance \cite{li2022human,moccia2018blood}. Beyond the standard Dice and Jaccard coefficients, researchers typically report pixel-wise precision and recall. Geometric measures such as the Hausdorff distance and average curve/surface distance are also employed \cite{schaap2009standardized,moccia2018blood}. To specifically address connectivity, metrics such as the percentage of correctly identified branches, total vessel length, and segment fragmentation are occasionally reported \cite{li2022human,moccia2018blood}, alongside topological measures like Betti numbers \cite{menten2023skeletonization}. Recently, the clDice metric \cite{shit2021cldice} has been introduced to quantify connectivity preservation by assessing the overlap between the skeletons of the ground truth and the prediction. 

A significant limitation of existing metrics is their focus on global accuracy, which often overlooks the varying difficulty levels inherent in vascular structures. For instance, in retinal fundus images, methods with high overall accuracy frequently underperform on thin vessels \cite{mookiah2021review}, which often exhibit low salience due to the finite resolution of the imaging device. Similarly, fluorescence microscopy can produce vessels with high contrast variability relative to the background (see Figure~\ref{f:examples}).

Recent studies \cite{reinke2021common,maier2024metrics} have highlighted the biases inherent in standard performance metrics, emphasizing that the application domain and the nature of the target object must be considered. We argue that variations in blood vessel salience must be accounted for during evaluation. If the vast majority of vessel segments in a sample are easily identifiable, a high recall score may merely indicate that an algorithm is capturing the most prominent structures while still failing on the more challenging, low-contrast segments.

The LVS index shares some conceptual similarities with vesselness metrics, such as the Frangi filter \cite{frangi1998multiscale}, which typically use the eigenvalues of the Hessian matrix \cite{lesage2009review,sato1998three}. However, the purpose and computation of the LVS index are distinct. Unlike vesselness filters, which are often used as a preprocessing step to enhance tubularity, the LVS index is an evaluative tool that requires ground-truth annotations. This reliance on manual annotation allows for a robust estimation of nearby vessel and background pixels. Furthermore, while vesselness filters often produce artifacts at bifurcations or terminations due to their non-tubular geometry, the LVS index is designed specifically to quantify segmentation quality across all vascular regions.

To our knowledge, no existing data augmentation method for neural network training systematically manipulates vessel salience. Standard practices include noise addition, blurring, random cropping, and color jittering. Elastic transformations \cite{da2022analysis,lin2022improving} are also commonly used to increase vessel tortuosity through random displacements. Our proposed augmentation fills this gap by specifically targeting the model's robustness to low-salience structures.

\section{Methodology}
\label{s:methodology}

\subsection{Local Vessel Salience}
\label{s:lvs}

We aim to characterize the local salience of blood vessel segments for every vascular pixel. We assume each image is accompanied by a ground-truth manual annotation mask. The first step involves representing this annotation as a graph, where nodes denote bifurcations and terminations, and edges represent vessel segments. For this task, we utilize the Pyvane framework\footnote{\url{https://github.com/chcomin/pyvane}}. Briefly, the framework applies a skeletonization algorithm to determine the medial axes of the vessels. Skeleton pixels with either a single neighbor (terminations) or three or more neighbors (bifurcations) are designated as graph nodes. A pruning procedure is applied to remove small, spurious branches, and a merging strategy consolidates adjacent nodes. Crucially, each graph edge contains the pixels of the medial axis for its respective vessel segment, hereafter referred to as the Medial Axis Segment (MAS). A complete description of this methodology is provided in~\cite{freitas2022unbiased}.

The objective is to calculate the salience at each MAS pixel and propagate these values to adjacent vascular pixels. The salience is calculated as the relative difference in intensity between the vessel and its surrounding background. More details are given below. For the calculation, it is first necessary to define the cross-section of the vessel at the position of the pixel. A possible approach is to define a normal vector with respect to the MAS. We found this approach unstable near sharp vessel turns, bifurcations, and terminations. Thus, we developed a simple and robust method that is guaranteed to find relevant vessel and background values for all MAS pixels. 

First, vessel boundaries are extracted from the ground-truth mask using a parametric contour-tracing algorithm \cite{suzuki1985topological}. For a given MAS pixel $p$, the closest contour pixel $p_{c1}$ is identified via Euclidean distance, and the direction vector $v_{c1} = p_{c1} - p$ is calculated. Next, the second closest contour pixel $p_{ci}$ is identified and a respective vector $v_{ci}=p_{ci}-p$ is defined. The dot product between vectors $v_{c1}$ and $v_{ci}$ is calculated. If it is negative, it means that the vectors have opposite directions, and thus two opposite contour points were found. A positive dot product means that $p_{ci}$ is at the same side of the contour as $p_{c1}$, and thus should be discarded. In such a case, the third closest contour pixel is identified and the dot product calculated, and so on until a contour pixel having a negative dot product with $v_{c1}$ is found. The final opposite pixel is represented as $p_{c2}$.

\begin{figure}
    \centering
    \includegraphics[width=0.8\linewidth]{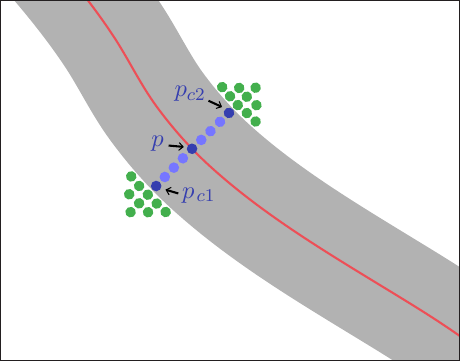}
    \caption{Illustration of the calculation of the LVS index. A blood vessel and the respective medial axis are represented, respectively, in gray and red. For a point $p$ in the medial axis, the two closest contour points $p_{c1}$ and $p_{c2}$ are shown. Pixels belonging to the set $S_v$ are shown in light and dark blue. Background pixels belonging to the set $S_b$ are shown in green.}
    \label{f:lvs_index}
\end{figure}

Given $p$, $p_{c1}$ and $p_{c2}$, two straight lines representing a cross-section of the vessel are defined. The first line, $l_1$, goes from $p_{c1}$ to $p$ while the second line, $l_2$, goes from $p$ to $p_{c2}$. Figure~\ref{f:lvs_index} shows an example of a MAS pixel and the respective shortest lines to nearby contour pixels. The pixels belonging to both lines are henceforth represented as $S_v$. The values of these pixels represent the local vessel signal that will be compared with the background.

Points $p_{c1}$ and $p_{c2}$ are used to identify nearby background pixels. All pixels with an Euclidean distance smaller than or equal to $r_b$ from $p_{c1}$ or from $p_{c2}$ are identified. The ground truth mask is used to discard pixels belonging to the blood vessel. Figure~\ref{f:lvs_index} shows an illustration of the background pixels identified at this step. These pixels are represented as $S_b$. 

Pixels $S_v$ and $S_b$ tend to form a dumbbell-like shape representing local intensities associated with $p$. $r_b$ is a free parameter of the method defining the scale where background pixels will be searched.

Different approaches can be used to calculate the local vessel salience from pixels $S_v$ and $S_b$. Hessian-based scores like the Frangi filter~\cite{frangi1998multiscale} could be used, but they often underestimate saliency at bifurcations and terminations by strictly assuming tubular geometry. We choose a score that is simple to calculate, has values in the range $[0,1]$, and can be easily interpreted. First, the average intensity vectors ${\bf I}_v$ and ${\bf I}_b$ of, respectively, $S_v$ and $S_b$ are calculated. Each vector has three values in the case of color images and a single value for grayscale images. Next, the relative difference in intensity is obtained as 

\begin{equation}
\Delta I=\frac{\Vert{\bf I}_v-{\bf I}_b\Vert}{\mathrm{max}(\Vert{\bf I}_v\Vert, \Vert{\bf I}_b\Vert)},\label{eq:lvs_nonorm}
\end{equation}
where $\Vert{\bf I}_x\Vert$ is the Euclidean norm of vector ${\bf I}_x$. 
$\Delta I$ can have sharp changes in value, which can hinder the definition of continuous regions for the recall measure presented in the next section. Thus, we smooth the values along the MAS. Since the pixels of the MAS are ordered parametrically along the segment, the final LVS index of a pixel $p_i$ is defined as

\begin{equation}
\mathrm{LVS}(p_i) = \frac{1}{2k+1}\sum_{j=i-k}^{i+k} \Delta I(p_j)\label{eq:lvs}
\end{equation}
where $\Delta I(p_j)$ represents the local difference for pixel $p_j$ of the MAS and parameter $k$ sets the degree of smoothing.

The final step of the calculation is to expand the values calculated for the MAS to other vessel pixels. For each vessel pixel, the closest MAS pixel is identified, and the respective LVS index is attributed to the pixel. The LVS index for each vessel pixel is stored as an image for use in downstream calculations.

\subsection{Low-Salience Recall}
\label{s:lsrecall}

Given the LVS index of each vessel pixel, it is possible to identify regions with different vessel salience. While pixels with high LVS values are generally easier to segment, those with low LVS present greater difficulty. Notably, a third class of pixels with zero LVS can be defined to represent vascular discontinuities. These pixels represent regions of vessel discontinuities in the image. Such pixels are very challenging to segment. Recent studies indicate that while humans exhibit a strong shape bias, CNNs rely more heavily on texture-based features~\cite{geirhos2018imagenettrained}. Thus, it is expected that humans perform better than CNNs at identifying vessel discontinuities, which can be located using the shape prior that two nearby and aligned, but discontinuous, segments are likely the same segment. Consequently, manual annotations may correctly identify these discontinuities as vascular structures, whereas CNNs often fail to recover such regions.

To evaluate the performance of a segmentation algorithm against a ground-truth reference, we define as $G$ the set of ground truth vessel pixels and as $R$ the set of vessel pixels identified by the algorithm. From these two sets, the number of true positive (TP) and false negative (FN) pixels can be calculated as

\begin{equation}
    TP = |G\cap R|
\end{equation}

\begin{equation}
    FN = |G-R|,
\end{equation}
where $|.|$ denotes the number of elements in the set. Then, the traditional recall metric can be defined as 

\begin{equation}
    \mathrm{Recall}=\frac{TP}{TP+FN}.
\end{equation}
Since the number of annotated blood vessel pixels is $P=TP+FN$, the recall metric quantifies the fraction of blood vessel pixels that are successfully identified by the algorithm. 

The LSRecall is defined using a similar calculation to the usual recall metric. Given the LVS index of all vessel pixels, a threshold value $t$ is defined, and only pixels having an LVS index equal to or smaller than the threshold are considered. This defines a binary image containing challenging vessel pixels that can be used in a usual recall calculation. We represent as $G_t$ the set of such pixels. It is important to note that $G_t$ is a subset of $G$. Then, the respective metrics are calculated as

\begin{equation}
    TP_t = |G_t\cap R|
\end{equation}

\begin{equation}
    FN_t = |G_t-R|
\end{equation}
The respective recall metric can be defined as 

\begin{equation}
    \mathrm{LSRecall}_t=\frac{TP_t}{TP_t+FN_t}.
\end{equation}

The LSRecall quantifies how successful the algorithm is at identifying challenging blood vessel pixels. Because $G_t$ may contain a limited number of pixels, significant fluctuations in LSRecall might result in only negligible changes to the overall recall metric. The threshold $t$ can be set to a specific value to quantify if a segmentation method is identifying pre-defined challenging regions or a LSRecall curve can be calculated using different values of $t$. 

To derive a single evaluation score independent of $t$, we propose a summary metric inspired by the standard practice in object detection of averaging Precision across multiple Intersection over Union (IoU) thresholds~\cite{lin2014microsoft}. The mean LSRecall, represented as mLSR, can be defined as the mean LSRecall for different $t$. Specifically, given a set of thresholds $T=\{t_b, t_b+0.01,t_b+0.02,\dots,t_f-0.01, t_f\}$, the respective $\mathrm{LSRecall}_t$ values are calculated for each threshold. $t_b$ can be automatically found as the smallest LVS value in a sample, ensuring that there are valid pixels for calculating the LSRecall. $t_f$ sets the largest LVS value to be considered in the calculation, it is defined as the maximum threshold at which LSRecall deviates from standard recall, marking the upper limit of salience for which segmentation remains challenging. The $\mathrm{mLSR}$ is then calculated as the mean of the $\mathrm{LSRecall}_t$ values for $t\in T$, that is,

\begin{equation}
\mathrm{mLSR} = \frac{1}{N}\sum_{t\in T}\mathrm{LSRecall}_t
\end{equation}
where $N$ is the number of thresholds used. 

An ideal algorithm that accurately segments vessels regardless of salience levels would achieve an mLSR of 1. Note that the $\mathrm{LSRecall}_t$ values are equally weighted in the average, even though they are calculated from different amount of ground truth pixels. The purpose is similar to the mean Average Precision metric~\cite{lin2014microsoft}: to heavily penalize errors on more challenging regions. 

Notably, like standard recall, the mLSR specifically assesses performance in relation to false negatives. Other metrics are required to quantify false positive rates.

\subsection{Augmenting Blood Vessel Salience}
\label{s:lsaug}

While this work primarily focuses on characterizing model performance in low-salience regions, we also evaluate a proof-of-concept approach designed to mitigate model bias toward low-salience vessels. This preliminary method is presented as a exploratory analysis to demonstrate feasibility. A comprehensive optimization of the procedure remains beyond the scope of this study.

Broadly, the salience of a subset of MAS pixels within a vessel segment is systematically reduced, with the option to introduce synthetic discontinuities. Specifically, starting from a designated point, the vessel intensity is gradually attenuated along the segment until it matches the local background intensity.

The spatial extent of the salience modification is governed by two parameters. Parameter $l$ sets the total salience modification length, that is, only a region of length $l$ of a segment is modified. Parameter $l_d<l$ sets the length of the discontinuity region. Thus, a region of length $l_d$ will have pixels with intensity identical to the local background intensity of the segment. A complete description of the augmentation methodology is presented in Appendix~\ref{app:aug}. Parameters $l$ and $l_d$ are randomly selected for each vessel segment during model training.

The augmentation creates vessels with varying degrees of salience. The hypothesis is that models trained on the augmented images will be less dependent on the local texture of the vessels and will make better use of continuity cues on segments with low salience.

\subsection{Datasets and segmentation methods employed}

We evaluated 16 segmentation methods across six blood vessel datasets using the mLSR metric. The DRIVE~\cite{StaalDRIVE}, VessMAP~\cite{silva2025new}, DCA1~\cite{cervantes2019automatic}, OCTA-500~\cite{li2024octa}, STARE~\cite{hoover2000locating} and CHASEDB1~\cite{fraz2012ensemble} datasets were employed. Relevant metadata for these datasets are summarized in Table~\ref{t:datasets}.

\begin{table}
    \centering
    \caption{Summary of the datasets used in the experiments.}
    \label{t:datasets}
    \resizebox{\columnwidth}{!}{%
    \begin{tabular}{@{}llll@{}}
        \toprule
        \textbf{Dataset} & \textbf{Modality} & \textbf{\# Images} & \textbf{Resolution} \\ \midrule
        DRIVE & Fundus Image & 40 & $565 \times 584$ \\
        STARE & Fundus Image & 20 & $700 \times 605$ \\
        CHASEDB1 & Fundus Image & 28 & $999 \times 960$ \\
        VessMAP & Fluorescence Microscopy & 100 & $256\times 256$ \\
        DCA1 & Coronary Angiography & 130 & $300 \times 300$ \\
        OCTA-500 & OCTA & 500 & $400 \times 400$ \\ \bottomrule
    \end{tabular}%
    }
\end{table}

The DRIVE dataset is a widely recognized benchmark in the literature. Comprising standard color fundus photographs, it frequently serves as the baseline for evaluating vessel segmentation algorithms. Most of the samples do not contain pathologies, but the manual annotations provided are highly consistent.

VessMAP, derived from fluorescence microscopy of cortical brain tissue, was developed to evaluate model robustness against out-of-distribution data, specifically focusing on the heterogeneity of vascular structures across varying noise and contrast levels. The DCA1 dataset is the prevalent benchmark for coronary angiography. The images were captured using X-ray during clinical procedures.

The OCTA-500 dataset consists of Optical Coherence Tomography Angiography samples. This modality allows for a detailed non-invasive look at the microvasculature. The 2D projections between the limiting membrane (ILM) and outer plexiform layer (OPL) (called projection B5 in the original paper~\cite{li2024octa}) were used. The $304\times 304$ images from the 3mm field of view were padded to $400\times 400$, which is the size of the 6mm field of view samples. Furthermore, we restricted our experiments to large vessel annotations, as small capillaries present significant annotation challenges and may introduce label noise.

\begin{table}
    \centering
    \caption{Overview of the methods employed in the experiments.}
    \label{t:method_comparison}
    \begin{tabularx}{\columnwidth}{@{}lcXc@{}}
        \toprule
        \textbf{Method} & \textbf{Year} & \textbf{Core Innovation} & \textbf{Ref.} \\ \midrule
UNet		& 2015 & Symmetrical encoder-decoder structure utilizing skip connections to preserve spatial information.			& \cite{ronneberger2015u}	\\ \addlinespace
UNet++		& 2018 & Nested and dense skip pathways designed to reduce the semantic gap between feature maps.					& \cite{zhou2018unet++}		\\ \addlinespace
ResUNet		& 2018 & Integration of residual units within the U-Net architecture to facilitate deeper training.					& \cite{zhang2018road}		\\ \addlinespace
AttUNet		& 2018 & Use of attention gates to automatically learn to focus on target structures of varying shapes.				& \cite{oktay2018attention}	\\ \addlinespace
R2UNet		& 2018 & Combination of recurrent convolutional layers and residual connections for iterative feature refinement.	& \cite{alom2018nuclei}		\\ \addlinespace
ResUNet++	& 2019 & Incorporation of ASPP, Squeeze-and-Excitation blocks, and attention gates.				& \cite{jha2019resunet++}	\\ \addlinespace
AGNet		& 2019 & Attention-guided filters specifically designed to suppress non-vascular noise in retinal images.			& \cite{zhang2019attention}	\\ \addlinespace
HRNet		& 2019 & Parallel multi-resolution streams that maintain high-resolution representations throughout the network.	& \cite{wang2020deep}		\\ \addlinespace
UNet3+		& 2020 & Full-scale skip connections and deep supervision for capturing multi-scale semantic information.			& \cite{huang2020unet}		\\ \addlinespace
SAUNet		& 2021 & Spatial attention module that prioritizes vessel-related pixels while suppressing background interference.	& \cite{guo2021sa}			\\ \addlinespace
ConvUNeXt	& 2022 & Modernization of the U-Net using ConvNeXt design principles like large kernels and layer normalization.	& \cite{han2022convunext}	\\ \addlinespace
FRUNet		& 2022 & A full-resolution stream that operates in parallel to avoid spatial detail loss for thin vessels.			& \cite{liu2022full}		\\ \addlinespace
LWUNet		& 2022 & A minimalistic, lightweight design of UNet optimized for state-of-the-art performance with low parameter counts.	& \cite{galdran2022state}	\\ \addlinespace
DCSAUNet	& 2023 & Compact architecture utilizing split-attention blocks for deeper and more efficient feature extraction.	& \cite{xu2023dcsau}		\\ \addlinespace
FSGNet		& 2025 & Full-scale representation and cross-scale guidance for integrating fine texture with global context.		& \cite{seo2025full}		\\ \addlinespace
FSGNet-b	& 2025 & The FSGNet model without the Guided Residual Modules for feature refinement.								& \cite{seo2025full}		\\ \bottomrule
    \end{tabularx}
\end{table}

Two other fundus photography datasets were used due to their popularity. The STARE dataset is valued for its clinical diversity. A significant portion of its images contains pathological features like hemorrhages and exudates, allowing to evaluate if a model can distinguish between blood vessels and disease-related lesions. The CHASEDB1 dataset contain images with uneven illumination and significant background noise.

Where available, official dataset splits are utilized. For datasets lacking a predefined split, a random 50/50 train-test allocation ism implemented, adhering to common practices for these benchmarks.

Table~\ref{t:method_comparison} provides an overview of the methods employed. Methods such as AGNet, SAUNet, FRUNet, LWUNet, and FSGNet are specialized architectures designed specifically for vascular segmentation. The remaining architectures, while originally developed for general medical image segmentation, are established benchmarks frequently applied to vascular tasks. Although they share similar objectives, they differ according to some design principles. For instance, UNet, UNet++, and UNet3+ represent evolutionary variations in information propagation between the encoder and decoder components. Architectures such as FRUNet, HRNet, and FSGNet prioritize the preservation of thin structures by avoiding aggressive downsampling, albeit at the cost of higher memory consumption. AttUNet, SAUNet and AGNet employ attention mechanisms for providing global context while preserving small details. More recent models, including LWUNet, ConvUNeXt, and DCSAUNet, focus on optimizing computational efficiency without compromising segmentation accuracy.

\section{Results}
\label{s:results}

For brevity, the LVS and LSRecall calculations are illustrated using the LWUNet model trained on the DRIVE (fundus) and VessMAP datasets. The mLSR metric was evaluated across all previously described methods and datasets. To improve readability, performance metrics are displayed in percentage. All experiments were conducted on a workstation equipped with an Intel Core i9 12900K processor, an NVIDIA RTX 3090 GPU, and 64 GB of RAM. The software implementation utilized PyTorch 2.10.0 and CUDA 12.9.

\subsection{Local Vessel Salience Analysis}

\begin{figure}
    \centering
    \includegraphics[width=0.75\linewidth]{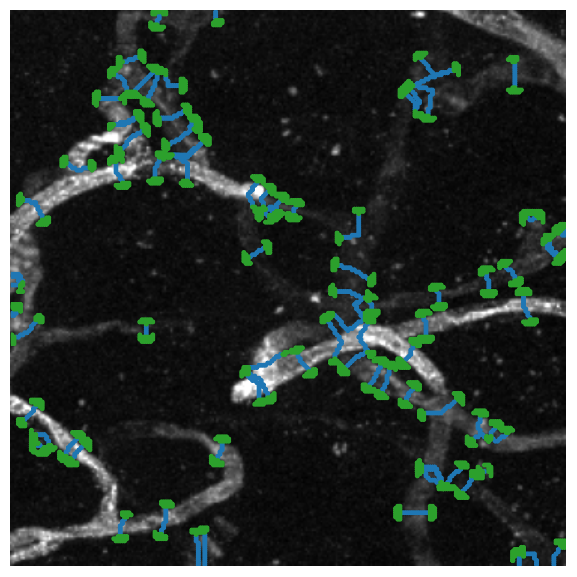}
    \caption{Examples of regions used for calculating the LVS in the VessMAP dataset. The image shows the vessel (blue) and background (green) points used for calculating the LVS of 100 randomly selected pixels of the medial axes of the vessels. Please see Figure~\ref{f:lvs_index} for the meaning of the dumbbell-like structures shown.}
    \label{f:lvs_sampling}
\end{figure}

\begin{figure*}
    \centering
    \includegraphics[width=\linewidth]{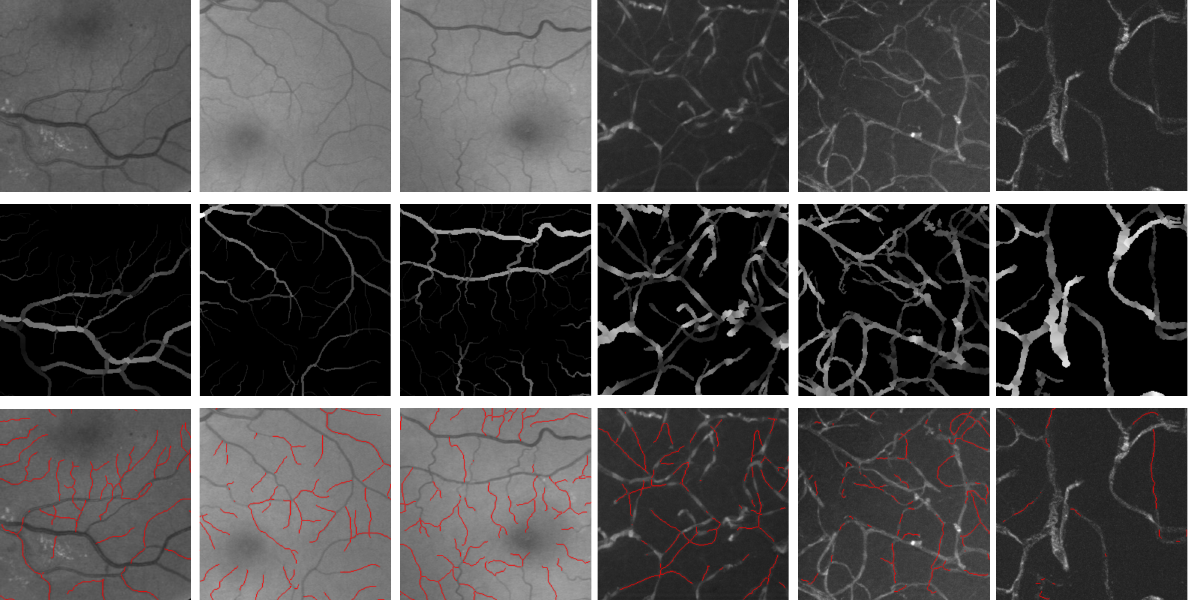}
    \caption{LVS index for some samples of the DRIVE (first three columns) and VessMAP (columns four to six) datasets. The first row of images shows the original samples, and the second row shows the LVS values calculated for each vessel pixel. The third row indicates in red regions where the LVS is smaller than 0.06 for the DRIVE dataset and 0.2 for the VessMAP dataset. The regions are indicated only at the medial axes of the blood vessels for easier comparison with the original samples.}
    \label{f:lvs_res}
\end{figure*}

To evaluate whether the LVS index effectively identifies challenging regions for segmentation, we calculated the LVS indices for all vascular pixels in the DRIVE and VessMAP datasets. For all experiments, a value of $r_b=4$ was used for identifying background pixels for the LVS calculation, and $k=15$ was used for smoothing the values (Equation~\ref{eq:lvs}). Section~\ref{s:sens} provides a sensitivity analysis of the results with respect to these parameters. Figure~\ref{f:lvs_sampling} shows examples of sampling regions used for calculating the LVS of some randomly selected vessel pixels for an image of the VessMAP dataset. We verified that for every medial axis pixel, the method successfully samples appropriate sets of vessel and background pixels to calculate the LVS index.

Figure~\ref{f:lvs_res} shows example results for three samples of the DRIVE and VessMAP datasets. In the fundus dataset, we observed that the LVS index generally correlates with vessel diameter (caliber), but it is possible to observe thicker vessels having low LVS, such as at the lower left corner of the first sample in Figure~\ref{f:lvs_res}, and thin vessels with relatively large LVS, such as the sinuous vessel at the upper left corner of the third sample. The Pearson correlation coefficient between the caliber of the vessels and respective LVS values was calculated for each sample and averaged among all samples of the DRIVE dataset. A value of 0.61 was obtained, suggesting that the LVS index and vessel diameter provide distinct information regarding vascular characteristics.

For the VessMAP dataset, the index tends to correlate with vessel intensity. However, the LVS index yields lower values in regions with higher background intensity or increased noise, as expected. A Pearson correlation coefficient of 0.43 was observed between the intensities of the pixels and respective LVS values.

\begin{figure}
    \centering
    \includegraphics[width=\linewidth]{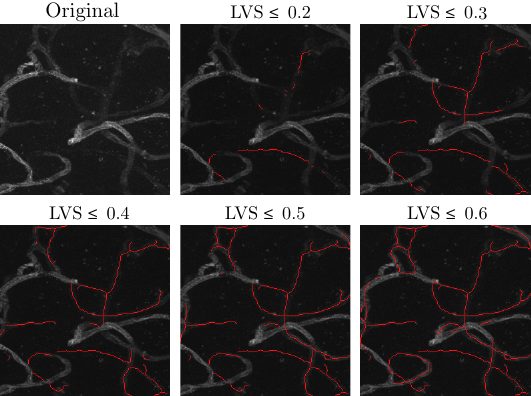}
    \caption{Blood vessels having different LVS values in a sample. Segments indicated in red have LVS lower than the threshold indicated above the respective image.}
    \label{f:lvs_thresh}
\end{figure}

Variable threshold values can be applied to identify specific challenging vascular regions for targeted performance evaluation. Figure~\ref{f:lvs_thresh} shows examples of thresholded indices using different threshold values. The threshold sets the expected difficulty in identifying vessel segments. Very low thresholds can be used to define regions where the blood vessels are almost indistinguishable from the background. In some cases, continuity criteria are required to identify regions connecting aligned yet discontinuous vessel segments.

In the sample shown in Figure~\ref{f:lvs_thresh}, segments having an LVS larger than 0.5 are easy to identify. Most segmentation methods should be able to identify these segments. Consequently, it is imperative to focus performance analysis on low-LVS vessels to gain more meaningful insights into algorithmic limitations.

\subsection{Quantifying Segmentation Quality Using LSRecall}
\label{s:lsrecall_res}

To evaluate the utility of the LSRecall metric in assessing performance within challenging vascular regions, we trained the LWUNet model~\cite{galdran2022state} on the DRIVE and VessMAP datasets, and the LSRecall metric was calculated on the results. The LWUNet model is a U-Net architecture with three downsampling and upsampling stages, each comprising two sets of convolution, batch normalization, and ReLU layers. The network was trained for 1,000 epochs with an initial learning rate of $0.01$, utilizing a polynomial decay scheduler (power of $0.9$) and a batch size of four. We employed the AdamW optimizer to minimize a cross-entropy loss function. The model achieving the highest Dice coefficient on the validation set was selected for final evaluation.

We utilized the official DRIVE dataset split ($20$ training and $20$ testing samples), partitioning the training set further into $16$ images for model optimization and $4$ for validation. For the VessMAP dataset, the train, validation, and test splits contained, respectively, 40, 10, and 50 samples. The data augmentation used on both datasets consisted of random rotation, translation, and scaling of the images together with random vertical and horizontal flips. 

The performance metrics obtained on the test set of the datasets are presented in Table~\ref{t:performance}. The resulting performance metrics are consistent with state-of-the-art values reported in~\cite{galdran2022state}.

\begin{table}[h]
    \centering
    \begin{tabular}{p{2.5cm}p{1.7cm}p{1.5cm}}

        \hline
        \textbf{Metric} & \textbf{DRIVE} & \textbf{VessMAP}        \\
        \hline
        AUC         & 97.8 & 98.6           \\
        Accuracy    & 95.3 & 94.9           \\
        Dice/F1     & 82.1 & 90.0           \\
        Specificity & 97.0 & 96.6            \\
        Recall      & 83.9 & 89.8            \\
        \hline 
    \end{tabular}
    \caption{Baseline performance metrics obtained for the LWUNet model on the test set of the DRIVE and VessMAP datasets. Values are displayed in percentage.}
    \label{t:performance}
\end{table}

\begin{figure}
    \centering
    \includegraphics[width=\linewidth]{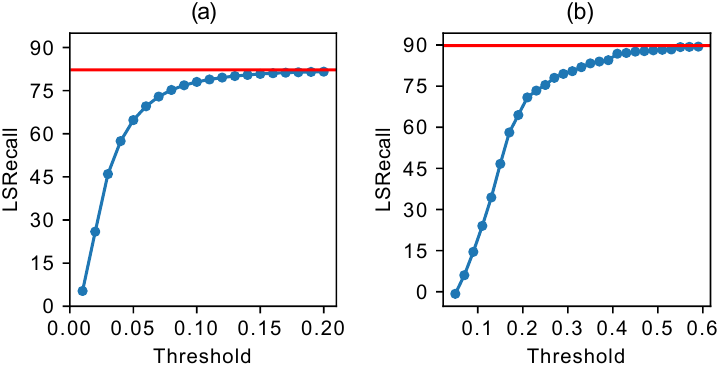}
    \caption{LSRecall as a function of the salience threshold used to calculate the metric for the (a) DRIVE and (b) VessMAP datasets. The red line shows the recall metric obtained for the dataset.}
    \label{f:lsrecall}
\end{figure}

LSRecall was computed for the test samples and averaged across each dataset. Figure~\ref{f:lsrecall} shows the values obtained for different threshold values used to calculate the LSRecall. The standard recall metric is indicated by a red horizontal line for comparison. Notably, lower vessel salience correlates with a systematic decrease in LSRecall, indicating that while the network effectively segments prominent vessels, it struggles to identify low-contrast structures.

Interestingly, performance degradation in VessMAP begins at an LVS of $0.4$, whereas DRIVE maintains higher accuracy until LVS values fall below $0.1$.

\subsection{Parameter Sensitivity Analysis}
\label{s:sens}

We proceed to the analysis of the mLSR score, which has two parameters: the background sampling radius $r_b$ and the smoothing window $k$. To evaluate their impact, we calculated mLSR values across the ranges $r_b \in [1, 13]$ and $k \in [0, 24]$ pixels. The upper bound for $r_b$ was selected to significantly exceed the average thickness of the vessels in the datasets, which is approximately 6 pixels. Consequently, $r_b=13$ samples a sufficient amount of background pixels (around 500 pixels) relative to the vessel pixels sampled, making further increases unnecessary. The upper bound for $k$ was determined by the average vessel segment length of 38 pixels measured in the datasets. Since $k=24$ defines a 49-pixel smoothing window, it typically encompasses the entire vessel segment, meaning larger values of $k$ should have no impact on the results.

\begin{figure}
    \centering
    \includegraphics[width=\linewidth]{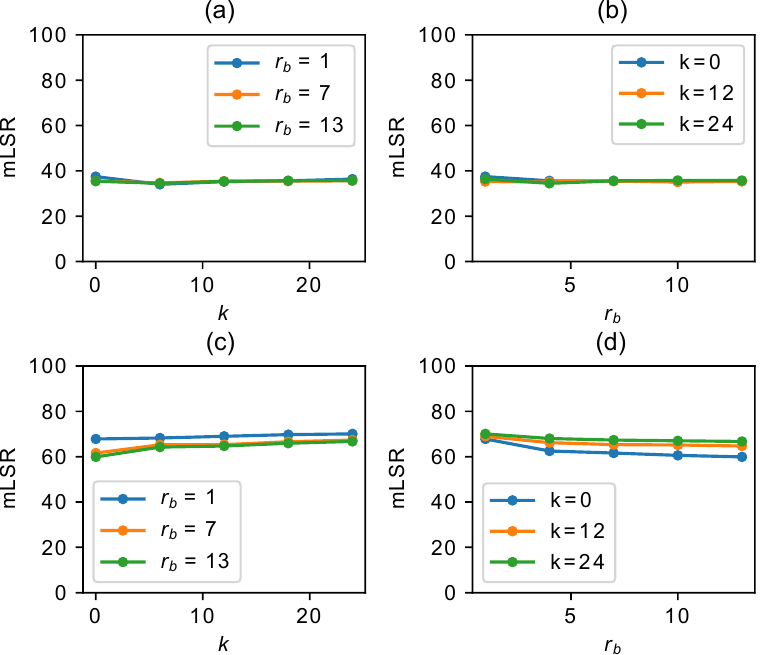}
    \caption{Parameter sensitivity analysis. The plots show the mLSR as a function of parameters $k$ and $r_b$ for the DRIVE (plots (a) and (b)) and VessMAP (plots (c) and (d)) datasets.}
    \label{f:perturbation}
\end{figure}

Figure~\ref{f:perturbation} shows the results for the DRIVE (Figures~\ref{f:perturbation}(a) and (b)) and VessMAP (Figures~\ref{f:perturbation}(c) and (d)) datasets. For the DRIVE dataset, the mLSR score remains largely invariant to the chosen parameter values. In the VessMAP dataset, minor fluctuations occur when $r_b$ and $k$ approach their minimum values ($r_b=1$, $k=0$). However, the metric demonstrates robustness to parameter variations for $k > 5$.

\subsection{Evaluation of Segmentation Methods}

The mLSR metric was used for evaluating the 16 methods described in Table~\ref{t:method_comparison}. To ensure convergence across all models regardless of size or inductive bias, we implemented a more robust training protocol than that described in previous sections. We employed the codebase from~\cite{seo2025full}\footnote{Available at \url{https://github.com/ZombaSY/FSG-Net-pytorch}} since it was recently used for the same purpose of comparing many different models. Model weights were optimized using AdamW with a weight decay of $0.05$ and a batch size of four. The learning rate followed a linear warmup from $0$ to $0.001$ over $20$ epochs, followed by a cosine scheduler that changes the learning rate from 0.001 to 0 and back to 0.001 over a 100-epoch cycle. The cycle is repeated until the Dice score on the validation set does not improve for 400 epochs (4 full cycles). Although this may lead to very long training times, it guarantees convergence of all models.

The optimization objective was defined as the unweighted sum of the Dice and Binary Cross-Entropy losses. Furthermore, deep supervision was applied since many models expect it. For models outputting multiple intermediate feature maps, the loss was calculated and averaged across all outputs. Images were padded to the next multiple of 32 of the largest size to avoid downsampling problems with the models. Specifically, an image with maximum dimension $n$ was padded to a size of $n_p \times n_p$, where $n_p = n + 32 - (n \bmod 32)$. $\bmod$ represents the remainder. 

A thorough data augmentation pipeline was applied. CutMix~\cite{yun2019cutmix} was applied to blend patches from randomly selected images, followed by random resizing between 0.5x and 2.0x scale and a random crop of size $288 \times 288$. Random horizontal flipping was used together with color jittering and random Gaussian blurring using a variable kernel size between 3 and 11 to add variability to the samples. 

Table~\ref{tab:metrics} shows the results averaged over the DRIVE, VessMAP, DCA1, OCTA-500, STARE and CHASEDB1 datasets. The full results and significance tests for each dataset are shown in Appendix~\ref{app:full}. The majority of the evaluated methods yielded comparable results. For instance, the top nine methods achieved Dice scores within one percentage point of each other. Interestingly, the original UNet model obtained the best performance. The metrics are highly correlated, indicating that they all correctly reflect the model's ability to segment the blood vessels.

\begin{table}
\caption{Segmentation performance of the 16 evaluated methods averaged over the 6 considered datasets, presented in descending order of Dice scores. All values are shown in percentage. AUC: Area Under the ROC curve. MCC: Mathews Correlation Coefficient. Acc: Accuracy. Prec: Precision. Rec: Recall.}
\label{tab:metrics}
\begin{tabular}{lp{0.81cm}p{0.81cm}p{0.81cm}p{0.81cm}p{0.81cm}p{0.81cm}p{0.9cm}}
\toprule
\textbf{Model} & \textbf{Dice} & \textbf{AUC} & \textbf{MCC} & \textbf{Acc} & \textbf{Prec} & \textbf{Rec} & \textbf{mLSR} \\
\midrule
UNet & 84.9 & 99.1 & 83.5 & 97.3 & 83.9 & 86.5 & 64.6 \\
AttUNet & 84.9 & 99.1 & 83.5 & 97.3 & 84.5 & 85.7 & 63.4 \\
ConvUNeXt & 84.8 & 99.1 & 83.5 & 97.3 & 84.5 & 85.7 & 63.5 \\
UNet3+ & 84.8 & 99.2 & 83.5 & 97.3 & 84.2 & 86.0 & 64.2 \\
FSGNet & 84.8 & 99.1 & 83.5 & 97.3 & 84.1 & 86.1 & 63.8 \\
HRNet-t & 84.6 & 99.1 & 83.2 & 97.3 & 83.8 & 85.9 & 64.5 \\
ResUNet & 84.5 & 99.1 & 83.1 & 97.3 & 83.9 & 85.7 & 63.8 \\
AGNet & 84.3 & 99.0 & 82.8 & 97.2 & 83.7 & 85.4 & 62.3 \\
FRUNet & 84.2 & 99.1 & 82.7 & 97.2 & 83.4 & 85.5 & 63.7 \\
SAUNet & 83.8 & 98.9 & 82.3 & 97.2 & 84.1 & 84.1 & 61.2 \\
DCSAUNet & 83.3 & 98.8 & 81.7 & 97.1 & 82.8 & 84.1 & 61.4 \\
FSGNet-b & 83.0 & 98.8 & 81.3 & 96.8 & 82.2 & 84.2 & 62.7 \\
ResUNet++ & 82.7 & 98.7 & 81.3 & 97.1 & 84.6 & 81.7 & 57.8 \\
LWUNet & 82.2 & 98.7 & 80.7 & 96.9 & 83.2 & 82.1 & 58.1 \\
R2UNet & 81.9 & 98.5 & 80.2 & 96.6 & 82.0 & 82.8 & 58.7 \\
UNet++ & 79.8 & 98.1 & 78.3 & 96.7 & 80.3 & 80.7 & 57.7 \\
\bottomrule
\end{tabular}
\end{table}

Focusing on the models' potential in recovering the blood vessels, Table~\ref{t:all_recalls} shows the recall and mLSR values obtained for each model on each dataset. The recall of the top performing methods is between 85.5\% and 86\%. On the other hand, the mLSR of the methods is significantly lower. We observed a typical decrease of $20$ percentage points in performance when specifically considering low-salience vessels. This indicates that all methods struggle to identify vessels with low LVS. The OCTA dataset seems to be the most challenging, even though only the large annotated vessels were considered for performance evaluation. It is closely followed by the DRIVE dataset. An interesting area for future investigation is determining whether this performance gap stems from architectural limitations or from the inherent difficulty and potential unreliability of human annotations for low-salience structures.

\begin{table}[htbp]
    \centering
    \caption{Comparison between the recall and mLSR obtained for each method on each dataset. The top value corresponds to the mLSR and the bottom gray value corresponds to the recall. Dataset names are abbreviated due to limited space. DRI: DRIVE. VMAP: VessMAP. CHA: CHASEDB1. STA: STARE.}
    \label{t:all_recalls}
    \begin{tabular}{lcccccc}
        \toprule
        \textbf{Model} & \textbf{DRI} & \textbf{VMAP} & \textbf{OCTA} & \textbf{DCA1} & \textbf{CHA} & \textbf{STA} \\
        \midrule
        UNet & \makecell{59.2 \\ \textcolor{gray}{83.8}} & \makecell{70.7 \\ \textcolor{gray}{91.0}} & \makecell{57.8 \\ \textcolor{gray}{89.9}} & \makecell{64.9 \\ \textcolor{gray}{83.1}} & \makecell{71.1 \\ \textcolor{gray}{86.2}} & \makecell{64.2 \\ \textcolor{gray}{84.8}} \\
        \addlinespace
        AttUNet & \makecell{58.9 \\ \textcolor{gray}{83.5}} & \makecell{65.5 \\ \textcolor{gray}{88.5}} & \makecell{57.7 \\ \textcolor{gray}{89.9}} & \makecell{62.3 \\ \textcolor{gray}{81.5}} & \makecell{70.2 \\ \textcolor{gray}{85.0}} & \makecell{65.8 \\ \textcolor{gray}{86.0}} \\
        \addlinespace
        ConvUNeXt & \makecell{57.7 \\ \textcolor{gray}{83.1}} & \makecell{66.7 \\ \textcolor{gray}{89.1}} & \makecell{57.4 \\ \textcolor{gray}{89.7}} & \makecell{63.0 \\ \textcolor{gray}{81.4}} & \makecell{70.2 \\ \textcolor{gray}{85.0}} & \makecell{65.7 \\ \textcolor{gray}{85.6}} \\
        \addlinespace
        UNet3+ & \makecell{58.6 \\ \textcolor{gray}{83.6}} & \makecell{67.1 \\ \textcolor{gray}{90.0}} & \makecell{57.6 \\ \textcolor{gray}{89.6}} & \makecell{64.1 \\ \textcolor{gray}{81.7}} & \makecell{72.1 \\ \textcolor{gray}{85.9}} & \makecell{65.7 \\ \textcolor{gray}{85.0}} \\
        \addlinespace
        FSGNet & \makecell{58.2 \\ \textcolor{gray}{83.6}} & \makecell{68.5 \\ \textcolor{gray}{89.6}} & \makecell{55.8 \\ \textcolor{gray}{89.3}} & \makecell{62.7 \\ \textcolor{gray}{82.3}} & \makecell{72.3 \\ \textcolor{gray}{85.7}} & \makecell{65.6 \\ \textcolor{gray}{85.9}} \\
        \addlinespace
        HRNet-t & \makecell{57.5 \\ \textcolor{gray}{82.6}} & \makecell{69.3 \\ \textcolor{gray}{89.9}} & \makecell{58.4 \\ \textcolor{gray}{90.1}} & \makecell{64.9 \\ \textcolor{gray}{82.5}} & \makecell{71.3 \\ \textcolor{gray}{84.8}} & \makecell{65.5 \\ \textcolor{gray}{85.6}} \\
        \addlinespace
        ResUNet & \makecell{58.5 \\ \textcolor{gray}{83.1}} & \makecell{68.2 \\ \textcolor{gray}{89.1}} & \makecell{56.9 \\ \textcolor{gray}{89.8}} & \makecell{63.4 \\ \textcolor{gray}{81.7}} & \makecell{69.1 \\ \textcolor{gray}{84.4}} & \makecell{66.8 \\ \textcolor{gray}{86.1}} \\
        \addlinespace
        AGNet & \makecell{55.6 \\ \textcolor{gray}{82.1}} & \makecell{69.5 \\ \textcolor{gray}{91.3}} & \makecell{54.7 \\ \textcolor{gray}{88.8}} & \makecell{62.0 \\ \textcolor{gray}{82.1}} & \makecell{69.5 \\ \textcolor{gray}{84.5}} & \makecell{62.2 \\ \textcolor{gray}{83.6}} \\
        \addlinespace
        FRUNet & \makecell{58.8 \\ \textcolor{gray}{83.5}} & \makecell{68.9 \\ \textcolor{gray}{89.4}} & \makecell{57.7 \\ \textcolor{gray}{90.1}} & \makecell{63.2 \\ \textcolor{gray}{81.8}} & \makecell{70.2 \\ \textcolor{gray}{84.8}} & \makecell{63.6 \\ \textcolor{gray}{83.4}} \\
        \addlinespace
        SAUNet & \makecell{57.4 \\ \textcolor{gray}{82.6}} & \makecell{68.0 \\ \textcolor{gray}{89.5}} & \makecell{56.1 \\ \textcolor{gray}{88.9}} & \makecell{60.6 \\ \textcolor{gray}{81.1}} & \makecell{65.2 \\ \textcolor{gray}{81.4}} & \makecell{59.9 \\ \textcolor{gray}{81.2}} \\
        \addlinespace
        DCSAUNet & \makecell{58.2 \\ \textcolor{gray}{83.7}} & \makecell{68.0 \\ \textcolor{gray}{90.4}} & \makecell{57.6 \\ \textcolor{gray}{90.2}} & \makecell{65.3 \\ \textcolor{gray}{82.9}} & \makecell{57.2 \\ \textcolor{gray}{75.0}} & \makecell{61.9 \\ \textcolor{gray}{82.5}} \\
        \addlinespace
        FSGNet-b & \makecell{59.5 \\ \textcolor{gray}{83.9}} & \makecell{65.4 \\ \textcolor{gray}{86.0}} & \makecell{58.2 \\ \textcolor{gray}{89.8}} & \makecell{61.5 \\ \textcolor{gray}{79.2}} & \makecell{69.6 \\ \textcolor{gray}{84.6}} & \makecell{62.1 \\ \textcolor{gray}{81.7}} \\
        \addlinespace
        ResUNet++ & \makecell{57.2 \\ \textcolor{gray}{82.6}} & \makecell{64.5 \\ \textcolor{gray}{88.1}} & \makecell{57.0 \\ \textcolor{gray}{89.8}} & \makecell{52.1 \\ \textcolor{gray}{73.3}} & \makecell{62.1 \\ \textcolor{gray}{79.6}} & \makecell{53.7 \\ \textcolor{gray}{76.5}} \\
        \addlinespace
        LWUNet & \makecell{56.1 \\ \textcolor{gray}{81.8}} & \makecell{62.8 \\ \textcolor{gray}{86.9}} & \makecell{52.9 \\ \textcolor{gray}{88.2}} & \makecell{57.9 \\ \textcolor{gray}{78.1}} & \makecell{63.1 \\ \textcolor{gray}{79.7}} & \makecell{55.7 \\ \textcolor{gray}{77.7}} \\
        \addlinespace
        R2UNet & \makecell{54.6 \\ \textcolor{gray}{81.7}} & \makecell{56.7 \\ \textcolor{gray}{81.8}} & \makecell{54.8 \\ \textcolor{gray}{88.0}} & \makecell{60.4 \\ \textcolor{gray}{80.8}} & \makecell{66.3 \\ \textcolor{gray}{82.2}} & \makecell{59.4 \\ \textcolor{gray}{82.1}} \\
        \addlinespace
        UNet++ & \makecell{53.4 \\ \textcolor{gray}{79.6}} & \makecell{68.8 \\ \textcolor{gray}{89.8}} & \makecell{58.4 \\ \textcolor{gray}{90.2}} & \makecell{51.9 \\ \textcolor{gray}{72.7}} & \makecell{66.3 \\ \textcolor{gray}{81.7}} & \makecell{47.5 \\ \textcolor{gray}{70.3}} \\
        \addlinespace
        \bottomrule
    \end{tabular}
\end{table}

\begin{figure}
    \centering
    \includegraphics[width=\linewidth]{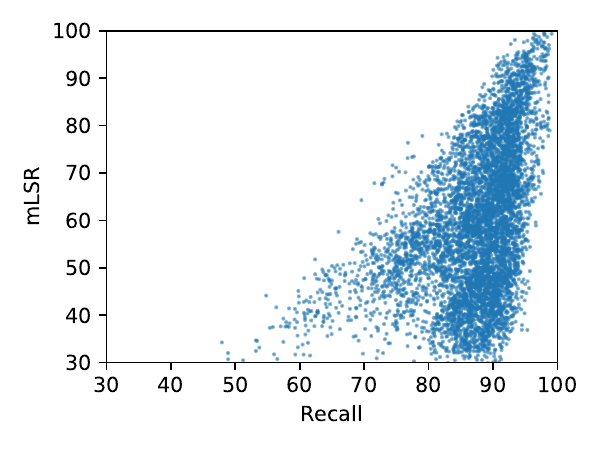}
    \caption{Recall and mLSR values obtained for all images and all methods evaluated in the experiments. Each point corresponds to the recall and mLSR obtained by a method applied to an image.}
    \label{f:rec_all}
\end{figure}

Recall and mLSR appear to be highly correlated. To further inspect the relationship between the two metrics, in Figure~\ref{f:rec_all} we show the values obtained for all methods on all images in the 6 datasets. At the image level, the two metrics diverge significantly, with a Pearson correlation coefficient of only $0.46$. The highest difference seems to be for images with high recall. Images with a recall of approximately $90\%$ can exhibit mLSR values as low as $30\%$, likely because high-salience vessels dominate the standard recall score.

\subsection{Salience Augmentation}
\label{sec:salience_augmentation}

\begin{figure*}
    \centering
    \includegraphics[width=\linewidth]{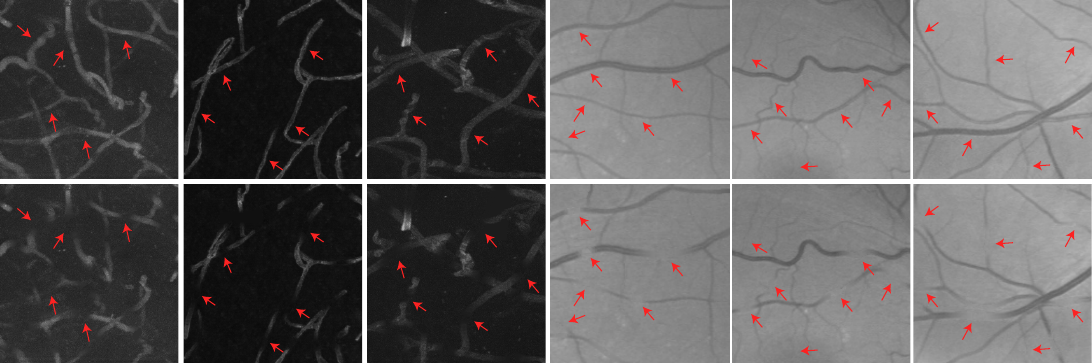}
    \caption{Examples of augmented blood vessels. The upper and lower rows of images show, respectively, the original and augmented samples. Red arrows indicate some augmented blood vessels. Low salience vessels are also augmented by the method but, as expected, they are difficult to visualize.}
    \label{f:lsrecall_aug}
\end{figure*}

We evaluated the proposed augmentation procedure to determine if models can develop increased robustness to variations in vascular salience. This preliminary inquiry was restricted to the VessMAP and DRIVE datasets to assess whether salience augmentation represents a promising research direction. The training protocol followed the methodology described in the previous section, with the addition of the salience augmentation step. For each training image, the number of augmented vessel segments, $n$, was randomly selected from a uniform distribution in the range $[50, 100]$. Figure~\ref{f:lsrecall_aug} provides visual examples of the resulting augmented vessels.

\begin{figure}
    \centering
    \includegraphics[width=\linewidth]{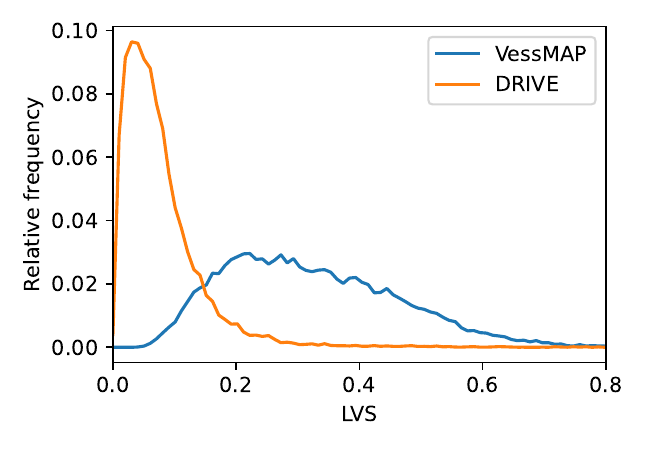}
    \caption{Histograms of the LVS values calculated for all samples of the DRIVE and VessMAP datasets.}
    \label{f:lvs_dist}
\end{figure}

For the DRIVE dataset, no improvement in performance metrics was observed. Similarly, the VessMAP dataset showed only marginal gains. For instance, the Dice score for the LWUNet model increased from $90.0\%$ to $90.7\%$. Furthermore, no significant increase in mLSR was recorded. The lack of improvement in the DRIVE dataset can be explained by the distribution of its LVS values, shown in Figure~\ref{f:lvs_dist}. The distribution exhibits a sharp peak at low LVS values, indicating that most vascular pixels already possess low salience. Consequently, the addition of synthetic salience degradation provides little additional benefit for this specific dataset.

As shown in Figure~\ref{f:lvs_dist}, the LVS values for the VessMAP dataset are more diverse than those in DRIVE. Given that low-salience vessels are inherently more difficult to detect and that the augmentation increases the prevalence of such pixels, a more significant performance improvement was anticipated. One hypothesis for the modest gain is that the original training set size was already sufficient to adequately represent the data distribution, even for challenging, hard-to-detect vessels.

To further explore the conditions under which the augmentation procedure enhances performance, we repeated the training using reduced training set sizes. Specifically, the LWUNet model was trained using only 2, 4, and 8 samples from the VessMAP dataset, with the remaining images reserved for testing. To account for variability between runs, each experiment was repeated 20 times using randomly selected training subsets. This analysis was omitted for the DRIVE dataset due to its consistently low and uniform LVS values. We utilized the LWUNet model exclusively for this experiment as its computational efficiency permitted the high number of training runs required for statistical validation.

The results of this analysis are summarized in Table~\ref{t:performance_aug}. Our findings indicate that the augmentation procedure significantly improves performance when training data is extremely scarce. For example, when training with only four samples, the Dice score improved from $75.3\%$ to $82.5\%$ upon applying salience augmentation, suggesting the method successfully enhanced the network's robustness to salience variations. Interestingly, the performance gain was more pronounced with four samples than with two. This is likely due to the inherent instability of models trained on only two samples, which resulted in poor segmentation quality across the test set. These results suggest that additional data augmentation strategies to prevent overfitting may be necessary to further improve performance at such minimal training set sizes.

\begin{table}[h]
    \centering
    \begin{tabular}{p{2.1cm}>{\centering\arraybackslash}p{2cm}>{\centering\arraybackslash}p{2cm}>{\centering\arraybackslash}p{2cm}}

        \hline
\textbf{Metric} & \textbf{2 samples} & \textbf{4 samples} & \textbf{8 samples}\\
\hline
AUC& 83.4$\pm$0.5& 91.2$\pm$0.6& 96.3$\pm$0.6\\
AUC (aug)& \cellcolor{gray!20}86.2$\pm$0.6& \cellcolor{gray!20}95.4$\pm$0.3& 96.4$\pm$0.6\\
\hline
Acc& 73.2$\pm$1.1& 85.0$\pm$0.8& 92.0$\pm$0.6\\
Acc (aug)& \cellcolor{gray!20}74.6$\pm$1.2& \cellcolor{gray!20}90.4$\pm$0.3& 91.9$\pm$0.7\\
\hline
Dice/F1& 61.6$\pm$0.8& 75.3$\pm$0.9& 85.7$\pm$0.7\\
Dice/F1 (aug)& \cellcolor{gray!20}64.7$\pm$1.0& \cellcolor{gray!20}82.5$\pm$0.5& 85.2$\pm$0.9\\
\hline
Spec.& 71.6$\pm$1.8& 85.9$\pm$1.2& 93.4$\pm$0.7\\
Spec. (aug)& \cellcolor{gray!20}72.9$\pm$1.8& \cellcolor{gray!20}92.1$\pm$0.4& 93.0$\pm$0.8\\
\hline
Recall& 77.7$\pm$1.2& 82.4$\pm$0.9& 87.9$\pm$0.5\\
Recall (aug)& \cellcolor{gray!20}82.2$\pm$1.0& \cellcolor{gray!20}85.6$\pm$0.7& 87.8$\pm$0.7\\
\hline
mLSR& 54.9$\pm$0.9& 54.0$\pm$1.4& 59.2$\pm$0.6\\
mLSR (aug)& \cellcolor{gray!20}57.7$\pm$0.8& \cellcolor{gray!20}58.3$\pm$1.1& \cellcolor{gray!20}61.2$\pm$0.5\\
\hline
    \end{tabular}
    \caption{Performance metrics obtained on the test of the VessMAP dataset set when training the LWUNet model with 2, 4, and 8 samples. The values represent the average and standard deviation calculated over 20 training runs. Metrics indicated with (aug) were calculated when salience augmentation was used. Shaded values indicate statistically significant improvements in the calculated averages according to a one-sided t-test with a p-value of 0.05. Values are displayed in percentage.}
    \label{t:performance_aug}
\end{table}

\section{Discussion}
\label{s:conclusion}

The proposed LVS index allows an intuitive quantification of the difficulty in segmenting blood vessels. Vessels with low LVS tend to be at the limit of a model's capacity. We introduced a variation of the recall metric, LSRecall, and a corresponding aggregate statistic, mLSR, to isolate and quantify segmentation quality within these challenging vessels. Our results demonstrate that segmentations achieving high global recall scores may still yield very low mLSR values. As expected, performance correlates with the LVS threshold used for the LSRecall calculation. Therefore, the LSRecall and mLSR metrics serve as valuable tools for characterizing systematic algorithmic biases across different vascular topologies. Furthermore, these metrics allow future development to be focused on high-difficulty regions, providing insights into how architectural designs and hyperparameter configurations impact performance.

The proposed data augmentation procedure yielded nuanced results in mitigating local errors driven by contrast variations. While no performance gains were observed for the DRIVE or VessMAP datasets when utilizing the full training sets, a statistically significant improvement was recorded with very small training sizes. This suggests that salience augmentation is a promising research direction for scenarios where annotated data is scarce, effectively helping the model generalize from limited examples.

A primary limitation of the LVS index is the assumption that salience can be quantified by simple intensity differences between the vessel and the surrounding tissue. In cases where vessel and non-vessel pixels possess similar intensities but distinct textures, the values derived from Equation~\ref{eq:lvs_nonorm} may not fully capture the difficulty. Still, a robust method was defined to sample sets of vessel and background pixels around each vessel point. This same framework could be adapted to incorporate more sophisticated measures of local difference beyond raw intensity.

The LVS index can be extended to 3D images using the same underlying methodology. For 3D volumes, only a portion of the vessel's cross-section would be sampled for the calculation, which should remain sufficient to provide a reliable estimation of local contrast. While an evaluation of 3D images was beyond the scope of this study since the 16 evaluated methods were designed for 2D data, we anticipate that the trends observed here would remain consistent for 3D vascular structures where salience is driven by intensity gradients.

Since all evaluated methods yielded similarly low mLSR values, it is important to question whether this stems from architectural inability to identify low-salience vessels or from inconsistencies in manual annotations. Low-LVS vessels are inherently challenging for human experts to identify and annotate. Thus, the performance of \emph{human annotators} is also expected to correlate with LVS. A compelling prospect for future research is to quantify the extent to which these performance gaps are due to annotation noise versus the inherent limitations of current models in representing subtle vascular appearances.

\section*{Funding}
C. H. Comin thanks FAPESP (grant no. 25/04800-9) for financial support. M. V. da Silva thanks FAPESP (grant no. 23/03975-4) and the Google PhD Fellowship Program for financial support.

\section*{CRediT Author Statement}
\textbf{João Pedro Parella}: Methodology, Software, Data curation. \textbf{Matheus Viana da Silva}: Software, Validation, Writing – review and editing. \textbf{Cesar Henrique Comin}: Conceptualization, Methodology, Supervision, Writing - original draft, review and editing.

\section*{Declaration of Competing Interest}
The authors declare that they have no known competing financial interests or personal relationships that could have appeared to influence the work reported in this paper.

\bibliography{references}

@article{li2022human,
  title={Human treelike tubular structure segmentation: A comprehensive review and future perspectives},
  author={Li, Hao and Tang, Zeyu and Nan, Yang and Yang, Guang},
  journal={Computers in Biology and Medicine},
  volume={151},
  pages={106241},
  year={2022},
  publisher={Elsevier}
}

@article{hu2021topology,
  title={Topology-aware segmentation using discrete morse theory},
  author={Hu, Xiaoling and Wang, Yusu and Fuxin, Li and Samaras, Dimitris and Chen, Chao},
  journal={arXiv preprint arXiv:2103.09992},
  year={2021}
}

@inproceedings{lin2014microsoft,
  title={Microsoft coco: Common objects in context},
  author={Lin, Tsung-Yi and Maire, Michael and Belongie, Serge and Hays, James and Perona, Pietro and Ramanan, Deva and Doll{\'a}r, Piotr and Zitnick, C Lawrence},
  booktitle={European conference on computer vision},
  pages={740--755},
  year={2014},
  organization={Springer}
}

@article{gupta2024topology,
  title={Topology-aware uncertainty for image segmentation},
  author={Gupta, Saumya and Zhang, Yikai and Hu, Xiaoling and Prasanna, Prateek and Chen, Chao},
  journal={Advances in Neural Information Processing Systems},
  volume={36},
  year={2024}
}

@article{sato1998three,
  title={Three-dimensional multi-scale line filter for segmentation and visualization of curvilinear structures in medical images},
  author={Sato, Yoshinobu and Nakajima, Shin and Shiraga, Nobuyuki and Atsumi, Hideki and Yoshida, Shigeyuki and Koller, Thomas and Gerig, Guido and Kikinis, Ron},
  journal={Medical image analysis},
  volume={2},
  number={2},
  pages={143--168},
  year={1998},
  publisher={Elsevier}
}

@article{lesage2009review,
  title={A review of 3D vessel lumen segmentation techniques: Models, features and extraction schemes},
  author={Lesage, David and Angelini, Elsa D and Bloch, Isabelle and Funka-Lea, Gareth},
  journal={Medical image analysis},
  volume={13},
  number={6},
  pages={819--845},
  year={2009},
  publisher={Elsevier}
}

@inproceedings{frangi1998multiscale,
  title={Multiscale vessel enhancement filtering},
  author={Frangi, Alejandro F and Niessen, Wiro J and Vincken, Koen L and Viergever, Max A},
  booktitle={Medical Image Computing and Computer-Assisted Intervention—MICCAI’98: First International Conference Cambridge, MA, USA, October 11--13, 1998 Proceedings 1},
  pages={130--137},
  year={1998},
  organization={Springer}
}

@article{moccia2018blood,
  title={Blood vessel segmentation algorithms—review of methods, datasets and evaluation metrics},
  author={Moccia, Sara and De Momi, Elena and El Hadji, Sara and Mattos, Leonardo S},
  journal={Computer methods and programs in biomedicine},
  volume={158},
  pages={71--91},
  year={2018},
  publisher={Elsevier}
}

@article{da2022analysis,
  title={An analysis of the influence of transfer learning when measuring the tortuosity of blood vessels},
  author={da Silva, Matheus V and Ouellette, Julie and Lacoste, Baptiste and Comin, Cesar H},
  journal={Computer Methods and Programs in Biomedicine},
  volume={225},
  pages={107021},
  year={2022},
  publisher={Elsevier}
}

@article{lin2022improving,
  title={Improving sensitivity and connectivity of retinal vessel segmentation via error discrimination network},
  author={Lin, Guoye and Bai, Hanhua and Zhao, Jie and Yun, Zhaoqiang and Chen, Yangfan and Pang, Shumao and Feng, Qianjin},
  journal={Medical Physics},
  volume={49},
  number={7},
  pages={4494--4507},
  year={2022},
  publisher={Wiley Online Library}
}

@article{fraz2012blood,
  title={Blood vessel segmentation methodologies in retinal images--a survey},
  author={Fraz, Muhammad Moazam and Remagnino, Paolo and Hoppe, Andreas and Uyyanonvara, Bunyarit and Rudnicka, Alicja R and Owen, Christopher G and Barman, Sarah A},
  journal={Computer methods and programs in biomedicine},
  volume={108},
  number={1},
  pages={407--433},
  year={2012},
  publisher={Elsevier}
}

@inproceedings{menten2023skeletonization,
  title={A skeletonization algorithm for gradient-based optimization},
  author={Menten, Martin J and Paetzold, Johannes C and Zimmer, Veronika A and Shit, Suprosanna and Ezhov, Ivan and Holland, Robbie and Probst, Monika and Schnabel, Julia A and Rueckert, Daniel},
  booktitle={Proceedings of the IEEE/CVF International Conference on Computer Vision},
  pages={21394--21403},
  year={2023}
}

@inproceedings{shit2021cldice,
  title={clDice-a novel topology-preserving loss function for tubular structure segmentation},
  author={Shit, Suprosanna and Paetzold, Johannes C and Sekuboyina, Anjany and Ezhov, Ivan and Unger, Alexander and Zhylka, Andrey and Pluim, Josien PW and Bauer, Ulrich and Menze, Bjoern H},
  booktitle={Proceedings of the IEEE/CVF conference on computer vision and pattern recognition},
  pages={16560--16569},
  year={2021}
}

@inproceedings{mosinska2018beyond,
  title={Beyond the pixel-wise loss for topology-aware delineation},
  author={Mosinska, Agata and Marquez-Neila, Pablo and Kozi{\'n}ski, Mateusz and Fua, Pascal},
  booktitle={Proceedings of the IEEE conference on computer vision and pattern recognition},
  pages={3136--3145},
  year={2018}
}

@article{funke2018large,
  title={Large scale image segmentation with structured loss based deep learning for connectome reconstruction},
  author={Funke, Jan and Tschopp, Fabian and Grisaitis, William and Sheridan, Arlo and Singh, Chandan and Saalfeld, Stephan and Turaga, Srinivas C},
  journal={IEEE transactions on pattern analysis and machine intelligence},
  volume={41},
  number={7},
  pages={1669--1680},
  year={2018},
  publisher={IEEE}
}

@article{scheffer2020connectome,
  title={A connectome and analysis of the adult Drosophila central brain},
  author={Scheffer, Louis K and Xu, C Shan and Januszewski, Michal and Lu, Zhiyuan and Takemura, Shin-ya and Hayworth, Kenneth J and Huang, Gary B and Shinomiya, Kazunori and Maitlin-Shepard, Jeremy and Berg, Stuart and others},
  journal={elife},
  volume={9},
  pages={e57443},
  year={2020},
  publisher={eLife Sciences Publications, Ltd}
}

@article{hu2019topology,
  title={Topology-preserving deep image segmentation},
  author={Hu, Xiaoling and Li, Fuxin and Samaras, Dimitris and Chen, Chao},
  journal={Advances in neural information processing systems},
  volume={32},
  year={2019}
}

@inproceedings{citraro2020towards,
  title={Towards reliable evaluation of algorithms for road network reconstruction from aerial images},
  author={Citraro, Leonardo and Kozi{\'n}ski, Mateusz and Fua, Pascal},
  booktitle={Computer Vision--ECCV 2020: 16th European Conference, Glasgow, UK, August 23--28, 2020, Proceedings, Part XXVIII 16},
  pages={703--719},
  year={2020},
  organization={Springer}
}

@inproceedings{vasu2020topoal,
  title={Topoal: An adversarial learning approach for topology-aware road segmentation},
  author={Vasu, Subeesh and Kozinski, Mateusz and Citraro, Leonardo and Fua, Pascal},
  booktitle={Computer Vision--ECCV 2020: 16th European Conference, Glasgow, UK, August 23--28, 2020, Proceedings, Part XXVII 16},
  pages={224--240},
  year={2020},
  organization={Springer}
}

@article{maier2024metrics,
  title={Metrics reloaded: recommendations for image analysis validation},
  author={Maier-Hein, Lena and Reinke, Annika and Godau, Patrick and Tizabi, Minu D and Buettner, Florian and Christodoulou, Evangelia and Glocker, Ben and Isensee, Fabian and Kleesiek, Jens and Kozubek, Michal and others},
  journal={Nature methods},
  pages={1--18},
  year={2024},
  publisher={Nature Publishing Group US New York}
}

@article{reinke2021common,
  title={Common limitations of image processing metrics: A picture story},
  author={Reinke, Annika and Tizabi, Minu D and Sudre, Carole H and Eisenmann, Matthias and R{\"a}dsch, Tim and Baumgartner, Michael and Acion, Laura and Antonelli, Michela and Arbel, Tal and Bakas, Spyridon and others},
  journal={arXiv preprint arXiv:2104.05642},
  year={2021}
}

@article{freitas2022unbiased,
  title={Unbiased analysis of mouse brain endothelial networks from two-or three-dimensional fluorescence images},
  author={Freitas-Andrade, Moises and Comin, Cesar H and da Silva, Matheus Viana and Costa, Luciano da F and Lacoste, Baptiste},
  journal={Neurophotonics},
  volume={9},
  number={3},
  pages={031916},
  year={2022},
  publisher={SPIE}
}

@inproceedings{paetzold2021whole,
  title={Whole brain vessel graphs: A dataset and benchmark for graph learning and neuroscience},
  author={Paetzold, Johannes C and McGinnis, Julian and Shit, Suprosanna and Ezhov, Ivan and B{\"u}schl, Paul and Prabhakar, Chinmay and Sekuboyina, Anjany and Todorov, Mihail and Kaissis, Georgios and Ert{\"u}rk, Ali and others},
  booktitle={Thirty-fifth Conference on Neural Information Processing Systems Datasets and Benchmarks Track (Round 2)},
  year={2021}
}

@article{suzuki1985topological,
  title={Topological structural analysis of digitized binary images by border following},
  author={Suzuki, Satoshi and others},
  journal={Computer vision, graphics, and image processing},
  volume={30},
  number={1},
  pages={32--46},
  year={1985},
  publisher={Elsevier}
}

@inproceedings{
geirhos2018imagenettrained,
title={ImageNet-trained {CNN}s are biased towards texture; increasing shape bias improves accuracy and robustness.},
author={Robert Geirhos and Patricia Rubisch and Claudio Michaelis and Matthias Bethge and Felix A. Wichmann and Wieland Brendel},
booktitle={International Conference on Learning Representations},
year={2019}
}

@article{mookiah2021review,
  title={A review of machine learning methods for retinal blood vessel segmentation and artery/vein classification},
  author={Mookiah, Muthu Rama Krishnan and Hogg, Stephen and MacGillivray, Tom J and Prathiba, Vijayaraghavan and Pradeepa, Rajendra and Mohan, Viswanathan and Anjana, Ranjit Mohan and Doney, Alexander S and Palmer, Colin NA and Trucco, Emanuele},
  journal={Medical Image Analysis},
  volume={68},
  pages={101905},
  year={2021},
  publisher={Elsevier}
}

@article{cervantes2023comprehensive,
  title={A comprehensive survey on segmentation techniques for retinal vessel segmentation},
  author={Cervantes, Jair and Cervantes, Jared and Garc{\'\i}a-Lamont, Farid and Yee-Rendon, Arturo and Cabrera, Josu{\'e} Espejel and Jalili, Laura Dom{\'\i}nguez},
  journal={Neurocomputing},
  volume={556},
  pages={126626},
  year={2023},
  publisher={Elsevier}
}

@article{chen2021retinal,
  title={Retinal vessel segmentation using deep learning: a review},
  author={Chen, Chunhui and Chuah, Joon Huang and Ali, Raza and Wang, Yizhou},
  journal={IEEE Access},
  volume={9},
  pages={111985--112004},
  year={2021},
  publisher={IEEE}
}

@inproceedings{van2008averaging,
  title={Averaging centerlines: mean shift on paths},
  author={van Walsum, Theo and Schaap, Michiel and Metz, Coert T and van der Giessen, Alina G and Niessen, Wiro J},
  booktitle={International Conference on Medical Image Computing and Computer-Assisted Intervention},
  pages={900--907},
  year={2008},
  organization={Springer}
}

@article{schaap2009standardized,
  title={Standardized evaluation methodology and reference database for evaluating coronary artery centerline extraction algorithms},
  author={Schaap, Michiel and Metz, Coert T and van Walsum, Theo and van der Giessen, Alina G and Weustink, Annick C and Mollet, Nico R and Bauer, Christian and Bogunovi{\'c}, Hrvoje and Castro, Carlos and Deng, Xiang and others},
  journal={Medical image analysis},
  volume={13},
  number={5},
  pages={701--714},
  year={2009},
  publisher={Elsevier}
}

@article{eladawi2018early,
  title={Early diabetic retinopathy diagnosis based on local retinal blood vessel analysis in optical coherence tomography angiography (OCTA) images},
  author={Eladawi, Nabila and Elmogy, Mohammed and Khalifa, Fahmi and Ghazal, Mohammed and Ghazi, Nicola and Aboelfetouh, Ahmed and Riad, Alaa and Sandhu, Harpal and Schaal, Shlomit and El-Baz, Ayman},
  journal={Medical physics},
  volume={45},
  number={10},
  pages={4582--4599},
  year={2018},
  publisher={Wiley Online Library}
}

@article{sangeethaa2018intelligent,
  title={An intelligent model for blood vessel segmentation in diagnosing {DR} using CNN},
  author={Sangeethaa, SN and Uma Maheswari, P},
  journal={Journal of medical systems},
  volume={42},
  number={10},
  pages={175},
  year={2018},
  publisher={Springer}
}

@article{li2021blood,
  title={Blood vessel segmentation of retinal image based on dense-U-Net network},
  author={Li, Zhenwei and Jia, Mengli and Yang, Xiaoli and Xu, Mengying},
  journal={Micromachines},
  volume={12},
  number={12},
  pages={1478},
  year={2021},
  publisher={MDPI}
}

@article{almotiri2018multi,
  title={A multi-anatomical retinal structure segmentation system for automatic eye screening using morphological adaptive fuzzy thresholding},
  author={Almotiri, Jasem and Elleithy, Khaled and Elleithy, Abdelrahman},
  journal={IEEE Journal of Translational Engineering in Health and Medicine},
  volume={6},
  pages={1--23},
  year={2018},
  publisher={IEEE}
}

@article{nair2020blood,
  title={Blood vessel segmentation and diabetic retinopathy recognition: an intelligent approach},
  author={Nair, Arun T and Muthuvel, K},
  journal={Computer Methods in Biomechanics and Biomedical Engineering: Imaging \& Visualization},
  volume={8},
  number={2},
  pages={169--181},
  year={2020},
  publisher={Taylor \& Francis}
}

@article{roda2021blood,
  title={Blood vessels and peripheral nerves as key players in cancer progression and therapy resistance},
  author={Roda, Niccol{\`o} and Blandano, Giada and Pelicci, Pier Giuseppe},
  journal={Cancers},
  volume={13},
  number={17},
  pages={4471},
  year={2021},
  publisher={MDPI}
}

@article{ouellette2020vascular,
  title={Vascular contributions to 16p11. 2 deletion autism syndrome modeled in mice},
  author={Ouellette, Julie and Toussay, Xavier and Comin, Cesar H and Costa, Luciano da F and Ho, Mirabelle and Lacalle-Aurioles, Mar{\'\i}a and Freitas-Andrade, Moises and Liu, Qing Yan and Leclerc, Sonia and Pan, Youlian and others},
  journal={Nature Neuroscience},
  volume={23},
  number={9},
  pages={1090--1101},
  year={2020},
  publisher={Nature Publishing Group US New York}
}

@article{dolati2015pre,
  title={Pre-operative image-based segmentation of the cranial nerves and blood vessels in microvascular decompression: can we prevent unnecessary explorations?},
  author={Dolati, Parviz and Golby, Alexandra and Eichberg, Daniel and Abolfotoh, Mohamad and Dunn, Ian F and Mukundan, Srinivasan and Hulou, Mohamed M and Al-Mefty, Ossama},
  journal={Clinical neurology and neurosurgery},
  volume={139},
  pages={159--165},
  year={2015},
  publisher={Elsevier}
}

@article{wong2019blood,
  title={Blood-brain barrier impairment and hypoperfusion are linked in cerebral small vessel disease},
  author={Wong, Sau May and Jansen, Jacobus FA and Zhang, C Eleana and Hoff, Erik I and Staals, Julie and van Oostenbrugge, Robert J and Backes, Walter H},
  journal={Neurology},
  volume={92},
  number={15},
  pages={e1669--e1677},
  year={2019},
  publisher={AAN Enterprises}
}

@inproceedings{yun2019cutmix,
  title={Cutmix: Regularization strategy to train strong classifiers with localizable features},
  author={Yun, Sangdoo and Han, Dongyoon and Oh, Seong Joon and Chun, Sanghyuk and Choe, Junsuk and Yoo, Youngjoon},
  booktitle={Proceedings of the IEEE/CVF international conference on computer vision},
  pages={6023--6032},
  year={2019}
}

@inproceedings{ronneberger2015u,
  title={U-net: Convolutional networks for biomedical image segmentation},
  author={Ronneberger, Olaf and Fischer, Philipp and Brox, Thomas},
  booktitle={International Conference on Medical image computing and computer-assisted intervention},
  pages={234--241},
  year={2015},
  organization={Springer}
}

@inproceedings{zhou2018unet++,
  title={Unet++: A nested u-net architecture for medical image segmentation},
  author={Zhou, Zongwei and Rahman Siddiquee, Md Mahfuzur and Tajbakhsh, Nima and Liang, Jianming},
  booktitle={International workshop on deep learning in medical image analysis},
  pages={3--11},
  year={2018},
  organization={Springer}
}

@article{zhang2018road,
  title={Road extraction by deep residual u-net},
  author={Zhang, Zhengxin and Liu, Qingjie and Wang, Yunhong},
  journal={IEEE Geoscience and Remote Sensing Letters},
  volume={15},
  number={5},
  pages={749--753},
  year={2018},
  publisher={IEEE}
}

@inproceedings{oktay2018attention,
title={Attention U-Net: Learning Where to Look for the Pancreas},
author={Ozan Oktay and Jo Schlemper and Loic Le Folgoc and Matthew Lee and Mattias Heinrich and Kazunari Misawa and Kensaku Mori and Steven McDonagh and Nils Y Hammerla and Bernhard Kainz and Ben Glocker and Daniel Rueckert},
booktitle={Medical Imaging with Deep Learning},
year={2018},
url={https://openreview.net/forum?id=Skft7cijM}
}

@inproceedings{alom2018nuclei,
  title={Nuclei segmentation with recurrent residual convolutional neural networks based U-Net (R2U-Net)},
  author={Alom, Md Zahangir and Yakopcic, Chris and Taha, Tarek M and Asari, Vijayan K},
  booktitle={NAECON 2018-IEEE National Aerospace and Electronics Conference},
  pages={228--233},
  year={2018},
  organization={IEEE}
}

@inproceedings{jha2019resunet++,
  title={Resunet++: An advanced architecture for medical image segmentation},
  author={Jha, Debesh and Smedsrud, Pia H and Riegler, Michael A and Johansen, Dag and De Lange, Thomas and Halvorsen, P{\aa}l and Johansen, H{\aa}vard D},
  booktitle={2019 IEEE international symposium on multimedia (ISM)},
  pages={225--2255},
  year={2019},
  organization={IEEE}
}

@inproceedings{zhang2019attention,
  title={Attention guided network for retinal image segmentation},
  author={Zhang, Shihao and Fu, Huazhu and Yan, Yuguang and Zhang, Yubing and Wu, Qingyao and Yang, Ming and Tan, Mingkui and Xu, Yanwu},
  booktitle={International conference on medical image computing and computer-assisted intervention},
  pages={797--805},
  year={2019},
  organization={Springer}
}

@article{wang2020deep,
  title={Deep high-resolution representation learning for visual recognition},
  author={Wang, Jingdong and Sun, Ke and Cheng, Tianheng and Jiang, Borui and Deng, Chaorui and Zhao, Yang and Liu, Dong and Mu, Yadong and Tan, Mingkui and Wang, Xinggang and others},
  journal={IEEE transactions on pattern analysis and machine intelligence},
  volume={43},
  number={10},
  pages={3349--3364},
  year={2020},
  publisher={IEEE}
}

@inproceedings{huang2020unet,
  title={Unet 3+: A full-scale connected unet for medical image segmentation},
  author={Huang, Huimin and Lin, Lanfen and Tong, Ruofeng and Hu, Hongjie and Zhang, Qiaowei and Iwamoto, Yutaro and Han, Xianhua and Chen, Yen-Wei and Wu, Jian},
  booktitle={ICASSP 2020-2020 IEEE international conference on acoustics, speech and signal processing (ICASSP)},
  pages={1055--1059},
  year={2020},
  organization={Ieee}
}

@inproceedings{guo2021sa,
  title={Sa-unet: Spatial attention u-net for retinal vessel segmentation},
  author={Guo, Changlu and Szemenyei, M{\'a}rton and Yi, Yugen and Wang, Wenle and Chen, Buer and Fan, Changqi},
  booktitle={2020 25th international conference on pattern recognition (ICPR)},
  pages={1236--1242},
  year={2021},
  organization={IEEE}
}

@article{han2022convunext,
  title={ConvUNeXt: An efficient convolution neural network for medical image segmentation},
  author={Han, Zhimeng and Jian, Muwei and Wang, Gai-Ge},
  journal={Knowledge-based systems},
  volume={253},
  pages={109512},
  year={2022},
  publisher={Elsevier}
}

@article{liu2022full,
  title={Full-resolution network and dual-threshold iteration for retinal vessel and coronary angiograph segmentation},
  author={Liu, Wentao and Yang, Huihua and Tian, Tong and Cao, Zhiwei and Pan, Xipeng and Xu, Weijin and Jin, Yang and Gao, Feng},
  journal={IEEE journal of biomedical and health informatics},
  volume={26},
  number={9},
  pages={4623--4634},
  year={2022},
  publisher={IEEE}
}

@article{galdran2022state,
  title={State-of-the-art retinal vessel segmentation with minimalistic models},
  author={Galdran, Adrian and Anjos, Andr{\'e} and Dolz, Jos{\'e} and Chakor, Hadi and Lombaert, Herv{\'e} and Ayed, Ismail Ben},
  journal={Scientific Reports},
  volume={12},
  number={1},
  pages={6174},
  year={2022},
  publisher={Nature Publishing Group UK London}
}

@article{xu2023dcsau,
  title={DCSAU-Net: A deeper and more compact split-attention U-Net for medical image segmentation},
  author={Xu, Qing and Ma, Zhicheng and Duan, Wenting and others},
  journal={Computers in biology and medicine},
  volume={154},
  pages={106626},
  year={2023},
  publisher={Elsevier}
}

@article{seo2025full,
  title={Full-scale representation guided network for retinal vessel segmentation},
  author={Seo, Sunyong and Yoo, Sangwook and Yoon, Huisu},
  journal={BMC Medical Imaging},
  volume={25},
  number={1},
  pages={484},
  year={2025},
  publisher={Springer}
}

@ARTICLE{StaalDRIVE,
  author={Staal, J. and Abramoff, M.D. and Niemeijer, M. and Viergever, M.A. and van Ginneken, B.},
  journal={IEEE Transactions on Medical Imaging}, 
  title={Ridge-based vessel segmentation in color images of the retina}, 
  year={2004},
  volume={23},
  number={4},
  pages={501-509}
}

@article{silva2025new,
  title={A new dataset for measuring the performance of blood vessel segmentation methods under distribution shifts},
  author={Silva, Matheus Viana da and de Carvalho Santos, Nat{\'a}lia and Ouellette, Julie and Lacoste, Baptiste and Comin, Cesar H},
  journal={PLOS ONE},
  volume={20},
  number={5},
  pages={1--25},
  year={2025},
  publisher={Public Library of Science}
}

@article{fraz2012ensemble,
  title={An ensemble classification-based approach applied to retinal blood vessel segmentation},
  author={Fraz, Muhammad Moazam and Remagnino, Paolo and Hoppe, Andreas and Uyyanonvara, Bunyarit and Rudnicka, Alicja R and Owen, Christopher G and Barman, Sarah A},
  journal={IEEE transactions on biomedical engineering},
  volume={59},
  number={9},
  pages={2538--2548},
  year={2012},
  publisher={IEEE}
}

@article{hoover2000locating,
  title={Locating blood vessels in retinal images by piecewise threshold probing of a matched filter response},
  author={Hoover, AD and Kouznetsova, Valentina and Goldbaum, Michael},
  journal={IEEE Transactions on Medical imaging},
  volume={19},
  number={3},
  pages={203--210},
  year={2000},
  publisher={IEEE}
}

@article{cervantes2019automatic,
  title={Automatic segmentation of coronary arteries in X-ray angiograms using multiscale analysis and artificial neural networks},
  author={Cervantes-Sanchez, Fernando and Cruz-Aceves, Ivan and Hernandez-Aguirre, Arturo and Hernandez-Gonzalez, Martha Alicia and Solorio-Meza, Sergio Eduardo},
  journal={Applied Sciences},
  volume={9},
  number={24},
  pages={5507},
  year={2019},
  publisher={MDPI}
}

@article{li2024octa,
  title={OCTA-500: a retinal dataset for optical coherence tomography angiography study},
  author={Li, Mingchao and Huang, Kun and Xu, Qiuzhuo and Yang, Jiadong and Zhang, Yuhan and Ji, Zexuan and Xie, Keren and Yuan, Songtao and Liu, Qinghuai and Chen, Qiang},
  journal={Medical image analysis},
  volume={93},
  pages={103092},
  year={2024},
  publisher={Elsevier}
}

\appendix
\section*{Appendices} 

\renewcommand{\thesection}{\Alph{section}}

\renewcommand{\thefigure}{\thesection\arabic{figure}}
\renewcommand{\theHfigure}{\thesection.\arabic{figure}}
\setcounter{figure}{0} 

\renewcommand{\thetable}{\thesection\arabic{table}}
\renewcommand{\theHtable}{\thesection.\arabic{table}}
\setcounter{table}{0} 

\section{Details of the Salience Augmentation Method}
\label{app:aug}

\begin{figure}[b]
    \centering
    \includegraphics[width=\linewidth]{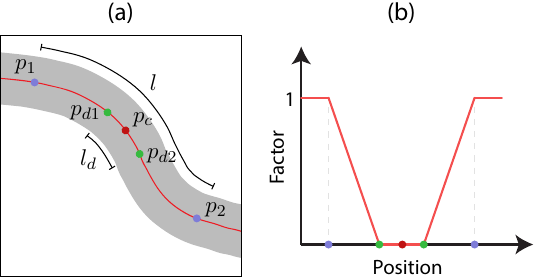}
    \caption{Illustration of the parameters involved in the salience augmentation. (a) The central point $p_c$ (red), initial and final points $p_1$ and $p_2$ (blue), and the points $p_{d1}$ and $p_{d2}$ defining the discontinuity region (green) are shown. Parameters $l$ and $l_d$ set the length of the salience augmentation and discontinuity region. (b) Intensity preservation factor along the MAS. Relevant positions are shown with the same colors as the points in (a).}
    \label{f:ls_aug}
\end{figure}

Given the medial axis (MAS) pixels of a specific blood vessel segment, a reference pixel $p_c$ is randomly selected. This pixel will be the center of the salience reduction region. From $p_c$, the two MAS pixels that are at a path-length distance of $l/2$ from $p_c$ are selected. These pixels, represented as $p_1$ and $p_2$, define the beginning and end of the salience reduction region. Two other pixels, $p_{d1}$ and $p_{d2}$, with path-length distance $l_d/2$ from $p_c$ are also identified. These pixels define the beginning and end of the discontinuity region that will have texture equal to the background. Figure~\ref{f:ls_aug}(a) illustrates the positions and lengths defined.

An intensity preservation factor $f_1$ is defined from $p_1$ to $p_{d1}$ as

\begin{equation}
    f_1(p_i) = \frac{d(p_i,p_{d1})}{d(p_1,p_{d1})}
\end{equation}
where $p_i$ represents a MAS pixel between $p_1$ and $p_{d1}$ and $d(p_x,p_y)$ represents the path-length distance between two points. Intuitively, the preservation factor is one for pixel $p_1$ and linearly decreases along the segment until reaching a value of zero at  $p_{d1}$. A preservation factor $f_2$ is similarly defined for pixels between $p_2$ and $p_{d2}$. Between pixels $p_{d1}$ and $p_{d2}$, the preservation factor is set to zero. A final preservation factor $f$ combining all calculated values is then defined. Figure~\ref{f:ls_aug}(b) shows the preservation factor for the MAS illustrated in Figure~\ref{f:ls_aug}(a).

The preservation factor is calculated only for the MAS pixels. Thus, they need to be expanded to the remaining vessel pixels. For each vessel pixel, the closest MAS pixel is identified (using the Euclidean distance) and the respective factor is associated with the vessel pixel. The final result is a preservation factor for every vessel pixel between points $p_1$ and $p_2$. A preservation factor of one is set to all other pixels of the vessel when necessary. 

Next, the intensities of the vessel segment are transformed. The minimum value needs to be similar to the local background intensity of the sample. This is done by taking the average value of all background pixels that are inside a rectangular region defined by the upper-left point $(p_c(x)-l,p_c(y)-l)$ and the lower-right point $(p_c(x)+l,p_c(y)+l)$ on image coordinates. $(p_c(x),p_c(y))$ are the coordinates of point $p_c$. The calculated value is represented as $I_m$. Next, the intensity $I(p)$ of each vessel pixel is transformed as

\begin{equation}
    \hat{I}(p) = f(I(p)-I_m) + I_m
\end{equation}
Thus, pixels close to $p_1$ and $p_2$ have little change in intensity, while pixels close to $p_{d1}$ and $p_{d2}$ become very similar to the background. Pixels between $p_{d1}$ and $p_{d2}$ become equal to $I_m$ since the preservation factor is zero in this region. 

A specific procedure was developed to make the discontinuous region as similar as possible to the background. Vessel pixels having a preservation factor of zero define a continuous image region $S_a$ that must have the appearance of the image background. A procedure was developed to replace the intensities at $S_a$ with a background region having the same shape as $S_a$. This was done by inverting the ground truth annotation and applying a binary erosion using $S_a$ as a structuring element with central point $p_a$. The remaining pixels are all centers of candidate background regions that can replace the intensities in $S_a$. 

\begin{figure}
    \centering
    \includegraphics[width=0.8\linewidth]{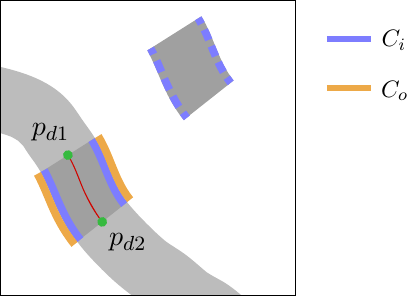}
    \caption{Illustration of the background identification method for creating a vessel discontinuity. The discontinuity region $S_a$ along the vessel is indicated in dark grey and in blue. The inner and outer contours $C_i$ and $C_o$ of $S_a$ are indicated, respectively, in blue and orange. A candidate background region is indicated in dark grey and as a dashed blue line. The pixel intensities $I_o$ are obtained from $C_o$ and the intensities $I_i$ are obtained from the pixels indicated in the dashed blue line.}
    \label{f:ls_aug2}
\end{figure}

The new background region might be significantly distinct from the local background of the discontinuous region $S_a$. Thus, a criterion was used to find good background candidates for replacement. Given $S_a$, the outer and inner contours of the set are obtained, as shown in Figure~\ref{f:ls_aug2}. The outer contour $C_o$ is defined as all background pixels having at least one 8-neighbor belonging to $S_a$. The inner contour $C_i$ is defined as all pixels in $S_a$ having at least one 8-neighbor in $C_o$. Thus, $C_o$ and $C_i$ define the pixels at the interface between the vessel and the background at the region where a vessel discontinuity is desired. The intensities $I_o$ of the outer contour are obtained and stored. The positions in $C_i$ are translated to a candidate background region found with the procedure described above, and the respective intensities $I_i$ are calculated. The average absolute difference between intensities of the outer and translated inner contour is obtained. The procedure is repeated to all candidate background regions.

A randomly selected background region having an absolute intensity difference smaller than a given threshold $t_b$ is identified, and the intensities are copied to the respective pixels of $S_a$. Thus, vessel pixels between $p_{d1}$ and $p_{d2}$ acquire new intensities copied from a background region having similar intensity to the local background intensity of pixels between $p_{d1}$ and $p_{d2}$. Threshold $t_b$ sets the largest acceptable difference between the local background and the background region selected to replace $S_a$.

For augmenting the images using the procedure described, the values $l$ and $l_d$ are randomly selected with uniform probability considering the constraint that $l$ must be smaller than the segment length and $l_d<l$. The number of vessel segments to be augmented in an image is also randomly selected.

\section{Full results}
\label{app:full}

\begin{figure*}
    \centering
    \includegraphics[width=\textwidth]{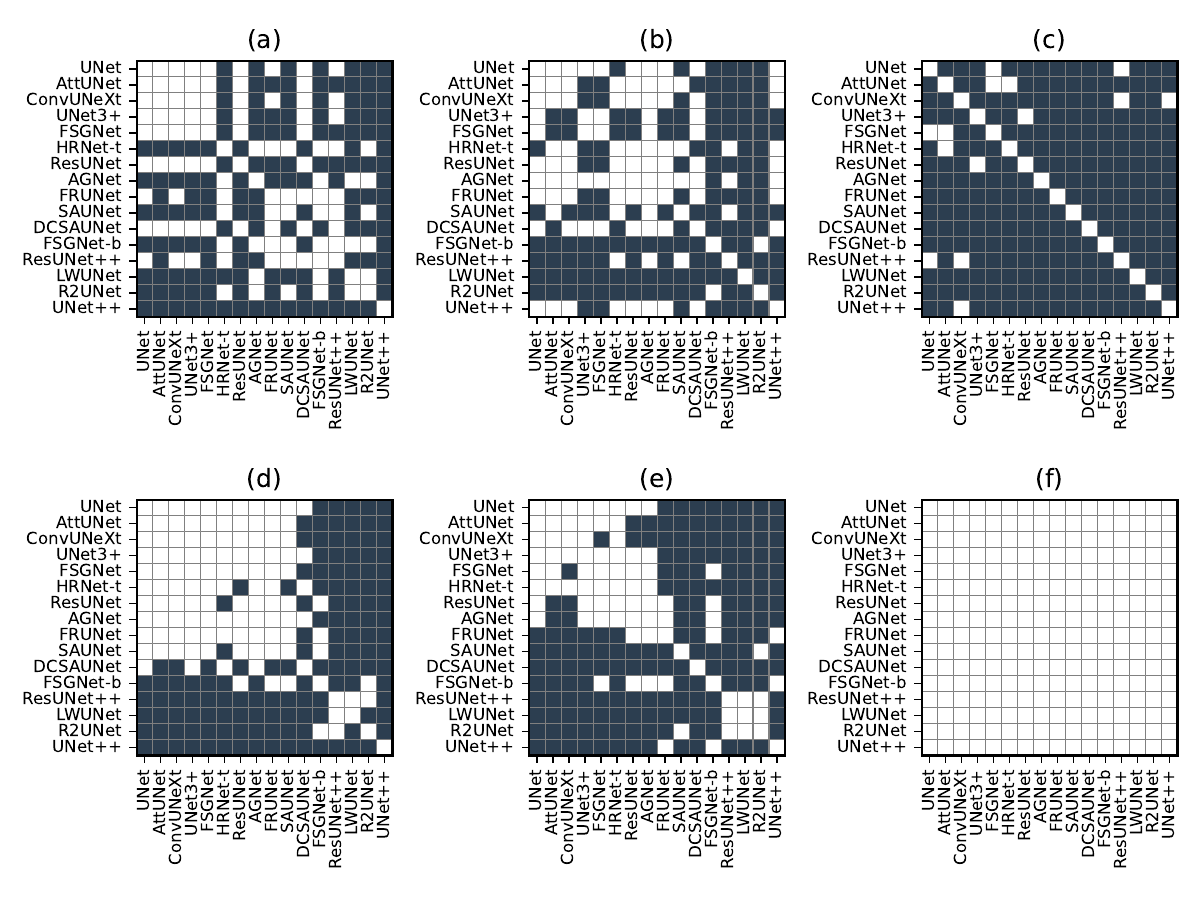}
    \caption{Pairwise statistical significance matrices comparing the 16 methods across 6 datasets. Each matrix represents a $16 \times 16$ pairwise comparison using the Wilcoxon signed-rank test on the Dice scores. P-values were adjusted using the Holm-Bonferroni correction to control for multiple comparisons ($N = 120$ comparisons per dataset). Dark cells indicate a statistically significant difference between methods ($p < 0.05$). (a) DRIVE, (b) VessMAP, (c) OCTA, (d) DCA1, (e) CHASEDB1, (f) STARE.}
    \label{f:significance}
\end{figure*}

Tables~\ref{t:full1} and~\ref{t:full2} show the segmentation performance metrics calculated for each method on the individual datasets considered in the experiments. Figure~\ref{f:significance} shows the result of the Wilcoxon signed-rank test between each pair of methods on each dataset.

\begin{table*}[htbp]
    \centering
    \footnotesize
    \caption{Segmentation performance of all methods evaluated on the DRIVE, VessMAP, OCTA and DCA1 datasets.}
    \label{t:full1}
    \begin{tabular}{l lp{1.cm}p{1.cm}p{1.cm}p{1.cm}p{1.cm}p{1.cm}p{1.cm}}
        \toprule
        \textbf{Dataset} & \textbf{Method} & \textbf{Dice} & \textbf{AUC} & \textbf{MCC} & \textbf{Acc} & \textbf{Prec} & \textbf{Rec} & \textbf{mLSR} \\
        \midrule
        \multirow{16}{*}{DRIVE} & UNet & 83.0 & 98.8 & 81.5 & 97.0 & 82.7 & 83.8 & 59.2 \\
         & AttUNet & 83.1 & 98.8 & 81.6 & 97.0 & 83.1 & 83.5 & 58.9 \\
         & ConvUNeXt & 82.9 & 98.8 & 81.4 & 97.0 & 83.2 & 83.1 & 57.7 \\
         & UNet3+ & 83.1 & 98.8 & 81.6 & 97.0 & 82.9 & 83.6 & 58.6 \\
         & FSGNet & 83.2 & 98.7 & 81.7 & 97.1 & 83.1 & 83.6 & 58.2 \\
         & HRNet-t & 82.5 & 98.7 & 80.9 & 96.9 & 82.8 & 82.6 & 57.5 \\
         & ResUNet & 83.1 & 98.8 & 81.6 & 97.0 & 83.4 & 83.1 & 58.5 \\
         & AGNet & 82.2 & 98.6 & 80.6 & 96.9 & 82.8 & 82.1 & 55.6 \\
         & FRUNet & 82.8 & 98.7 & 81.2 & 97.0 & 82.4 & 83.5 & 58.8 \\
         & SAUNet & 82.6 & 98.7 & 81.1 & 97.0 & 83.0 & 82.6 & 57.4 \\
         & DCSAUNet & 83.1 & 98.8 & 81.6 & 97.0 & 82.9 & 83.7 & 58.2 \\
         & FSGNet-b & 82.5 & 98.7 & 80.9 & 96.9 & 81.5 & 83.9 & 59.5 \\
         & ResUNet++ & 82.8 & 98.7 & 81.3 & 97.0 & 83.4 & 82.6 & 57.2 \\
         & LWUNet & 82.0 & 98.5 & 80.4 & 96.9 & 82.7 & 81.8 & 56.1 \\
         & R2UNet & 82.2 & 98.6 & 80.6 & 96.9 & 83.1 & 81.7 & 54.6 \\
         & UNet++ & 80.9 & 98.2 & 79.3 & 96.7 & 82.6 & 79.6 & 53.4 \\
        \midrule
        \multirow{16}{*}{VessMAP} & UNet & 91.5 & 99.1 & 88.7 & 95.8 & 92.3 & 91.0 & 70.7 \\
         & AttUNet & 91.1 & 99.1 & 88.3 & 95.7 & 94.1 & 88.5 & 65.5 \\
         & ConvUNeXt & 91.2 & 99.1 & 88.5 & 95.7 & 93.9 & 89.1 & 66.7 \\
         & UNet3+ & 91.6 & 99.2 & 89.0 & 95.9 & 93.6 & 90.0 & 67.1 \\
         & FSGNet & 91.7 & 99.2 & 89.1 & 96.0 & 94.3 & 89.6 & 68.5 \\
         & HRNet-t & 91.1 & 99.1 & 88.3 & 95.6 & 92.8 & 89.9 & 69.3 \\
         & ResUNet & 91.3 & 99.1 & 88.6 & 95.8 & 93.9 & 89.1 & 68.2 \\
         & AGNet & 91.3 & 99.1 & 88.5 & 95.7 & 91.7 & 91.3 & 69.5 \\
         & FRUNet & 91.2 & 99.1 & 88.4 & 95.7 & 93.4 & 89.4 & 68.9 \\
         & SAUNet & 90.8 & 99.0 & 87.9 & 95.5 & 92.5 & 89.5 & 68.0 \\
         & DCSAUNet & 91.5 & 99.1 & 88.8 & 95.8 & 93.0 & 90.4 & 68.0 \\
         & FSGNet-b & 87.5 & 98.2 & 83.4 & 93.8 & 89.2 & 86.0 & 65.4 \\
         & ResUNet++ & 90.7 & 99.0 & 87.8 & 95.5 & 93.7 & 88.1 & 64.5 \\
         & LWUNet & 89.3 & 98.7 & 86.1 & 94.8 & 92.5 & 86.9 & 62.8 \\
         & R2UNet & 85.6 & 97.2 & 81.8 & 93.1 & 91.5 & 81.8 & 56.7 \\
         & UNet++ & 91.3 & 99.1 & 88.5 & 95.7 & 93.2 & 89.8 & 68.8 \\
        \midrule
        \multirow{16}{*}{OCTA} & UNet & 89.9 & 99.6 & 89.1 & 98.4 & 90.1 & 89.9 & 57.8 \\
         & AttUNet & 90.0 & 99.6 & 89.2 & 98.4 & 90.3 & 89.9 & 57.7 \\
         & ConvUNeXt & 89.9 & 99.6 & 89.0 & 98.4 & 90.2 & 89.7 & 57.4 \\
         & UNet3+ & 89.6 & 99.6 & 88.7 & 98.3 & 89.8 & 89.6 & 57.6 \\
         & FSGNet & 90.0 & 99.6 & 89.1 & 98.4 & 90.8 & 89.3 & 55.8 \\
         & HRNet-t & 90.1 & 99.6 & 89.2 & 98.4 & 90.2 & 90.1 & 58.4 \\
         & ResUNet & 89.6 & 99.6 & 88.7 & 98.3 & 89.6 & 89.8 & 56.9 \\
         & AGNet & 88.6 & 99.4 & 87.6 & 98.1 & 88.5 & 88.8 & 54.7 \\
         & FRUNet & 89.3 & 99.6 & 88.4 & 98.2 & 88.7 & 90.1 & 57.7 \\
         & SAUNet & 88.7 & 99.5 & 87.8 & 98.2 & 88.8 & 88.9 & 56.1 \\
         & DCSAUNet & 90.1 & 99.6 & 89.3 & 98.4 & 90.3 & 90.2 & 57.6 \\
         & FSGNet-b & 88.9 & 99.5 & 87.9 & 98.2 & 88.1 & 89.8 & 58.2 \\
         & ResUNet++ & 89.9 & 99.6 & 89.0 & 98.4 & 90.1 & 89.8 & 57.0 \\
         & LWUNet & 88.1 & 99.5 & 87.2 & 98.1 & 88.4 & 88.2 & 52.9 \\
         & R2UNet & 87.1 & 99.3 & 86.1 & 97.9 & 86.5 & 88.0 & 54.8 \\
         & UNet++ & 89.8 & 99.6 & 88.9 & 98.3 & 89.6 & 90.2 & 58.4 \\
        \midrule
        \multirow{16}{*}{DCA1} & UNet & 79.3 & 99.1 & 78.4 & 97.6 & 76.9 & 83.1 & 64.9 \\
         & AttUNet & 79.0 & 99.1 & 78.1 & 97.6 & 77.9 & 81.5 & 62.3 \\
         & ConvUNeXt & 79.2 & 99.1 & 78.2 & 97.6 & 78.0 & 81.4 & 63.0 \\
         & UNet3+ & 79.2 & 99.1 & 78.3 & 97.6 & 77.9 & 81.7 & 64.1 \\
         & FSGNet & 79.2 & 99.1 & 78.3 & 97.6 & 77.3 & 82.3 & 62.7 \\
         & HRNet-t & 79.4 & 99.1 & 78.5 & 97.6 & 77.4 & 82.5 & 64.9 \\
         & ResUNet & 78.7 & 99.1 & 77.8 & 97.6 & 77.1 & 81.7 & 63.4 \\
         & AGNet & 79.6 & 99.1 & 78.6 & 97.7 & 78.1 & 82.1 & 62.0 \\
         & FRUNet & 78.7 & 99.1 & 77.8 & 97.6 & 77.1 & 81.8 & 63.2 \\
         & SAUNet & 78.5 & 99.1 & 77.7 & 97.6 & 77.6 & 81.1 & 60.6 \\
         & DCSAUNet & 79.9 & 99.2 & 78.9 & 97.7 & 77.9 & 82.9 & 65.3 \\
         & FSGNet-b & 78.2 & 99.0 & 77.2 & 97.5 & 78.3 & 79.2 & 61.5 \\
         & ResUNet++ & 75.6 & 98.8 & 74.8 & 97.4 & 79.9 & 73.3 & 52.1 \\
         & LWUNet & 75.7 & 98.8 & 74.9 & 97.3 & 75.7 & 78.1 & 57.9 \\
         & R2UNet & 77.3 & 98.8 & 76.4 & 97.4 & 75.4 & 80.8 & 60.4 \\
         & UNet++ & 64.5 & 96.8 & 63.6 & 95.6 & 61.4 & 72.7 & 51.9 \\
        \bottomrule
    \end{tabular}
\end{table*}

\begin{table*}[htbp]
    \centering
    \footnotesize
    \caption{Segmentation performance of all methods evaluated on the CHASEDB1 and STARE datasets.}
    \label{t:full2}
    \begin{tabular}{l lp{1.cm}p{1.cm}p{1.cm}p{1.cm}p{1.cm}p{1.cm}p{1.cm}}
        \toprule
        \textbf{Dataset} & \textbf{Method} & \textbf{Dice} & \textbf{AUC} & \textbf{MCC} & \textbf{Acc} & \textbf{Prec} & \textbf{Rec} & \textbf{mLSR} \\
        \midrule
        \multirow{16}{*}{CHASEDB1} & UNet & 80.6 & 99.0 & 79.5 & 97.5 & 75.9 & 86.2 & 71.1 \\
         & AttUNet & 80.9 & 98.9 & 79.7 & 97.5 & 77.3 & 85.0 & 70.2 \\
         & ConvUNeXt & 81.0 & 99.0 & 79.9 & 97.6 & 77.6 & 85.0 & 70.2 \\
         & UNet3+ & 80.5 & 98.9 & 79.4 & 97.4 & 75.9 & 85.9 & 72.1 \\
         & FSGNet & 80.3 & 98.9 & 79.1 & 97.4 & 75.6 & 85.7 & 72.3 \\
         & HRNet-t & 80.6 & 98.9 & 79.4 & 97.5 & 76.9 & 84.8 & 71.3 \\
         & ResUNet & 79.8 & 98.8 & 78.6 & 97.4 & 76.0 & 84.4 & 69.1 \\
         & AGNet & 80.0 & 98.7 & 78.7 & 97.4 & 76.0 & 84.5 & 69.5 \\
         & FRUNet & 79.3 & 98.7 & 78.1 & 97.3 & 74.6 & 84.8 & 70.2 \\
         & SAUNet & 78.2 & 98.1 & 76.8 & 97.2 & 75.5 & 81.4 & 65.2 \\
         & DCSAUNet & 72.9 & 97.4 & 71.1 & 96.6 & 71.2 & 75.0 & 57.2 \\
         & FSGNet-b & 79.6 & 98.8 & 78.4 & 97.3 & 75.3 & 84.6 & 69.6 \\
         & ResUNet++ & 76.7 & 97.8 & 75.2 & 97.0 & 74.3 & 79.6 & 62.1 \\
         & LWUNet & 76.6 & 98.1 & 75.2 & 97.0 & 74.2 & 79.7 & 63.1 \\
         & R2UNet & 77.5 & 98.3 & 76.1 & 97.1 & 73.5 & 82.2 & 66.3 \\
         & UNet++ & 79.0 & 98.7 & 77.7 & 97.3 & 76.7 & 81.7 & 66.3 \\
        \midrule
        \multirow{16}{*}{STARE} & UNet & 85.0 & 99.2 & 83.9 & 97.8 & 85.4 & 84.8 & 64.2 \\
         & AttUNet & 85.1 & 99.2 & 83.9 & 97.8 & 84.4 & 86.0 & 65.8 \\
         & ConvUNeXt & 84.8 & 99.2 & 83.7 & 97.7 & 84.4 & 85.6 & 65.7 \\
         & UNet3+ & 85.0 & 99.3 & 83.9 & 97.8 & 85.4 & 85.0 & 65.7 \\
         & FSGNet & 84.6 & 99.2 & 83.4 & 97.7 & 83.5 & 85.9 & 65.6 \\
         & HRNet-t & 83.9 & 99.1 & 82.6 & 97.5 & 82.5 & 85.6 & 65.5 \\
         & ResUNet & 84.7 & 99.2 & 83.5 & 97.7 & 83.6 & 86.1 & 66.8 \\
         & AGNet & 84.2 & 99.1 & 83.0 & 97.6 & 85.0 & 83.6 & 62.2 \\
         & FRUNet & 83.7 & 99.1 & 82.5 & 97.6 & 84.4 & 83.4 & 63.6 \\
         & SAUNet & 83.8 & 99.0 & 82.8 & 97.7 & 87.2 & 81.2 & 59.9 \\
         & DCSAUNet & 82.1 & 98.5 & 80.7 & 97.3 & 82.0 & 82.5 & 61.9 \\
         & FSGNet-b & 81.2 & 98.8 & 79.7 & 97.1 & 80.8 & 81.7 & 62.1 \\
         & ResUNet++ & 80.6 & 98.5 & 79.5 & 97.3 & 86.1 & 76.5 & 53.7 \\
         & LWUNet & 81.3 & 98.4 & 80.2 & 97.4 & 85.8 & 77.7 & 55.7 \\
         & R2UNet & 81.7 & 98.6 & 80.3 & 97.3 & 81.7 & 82.1 & 59.4 \\
         & UNet++ & 73.3 & 96.4 & 71.9 & 96.2 & 78.4 & 70.3 & 47.5 \\
        \bottomrule
    \end{tabular}
\end{table*}

\end{document}